\documentclass[pdflatex,sn-mathphys-num]{sn-jnl}% Math and Physical Sciences Numbered Reference Style

\usepackage{lineno}
\usepackage{graphicx}%
\usepackage{multirow}%
\usepackage{amsmath,amssymb,amsfonts}%
\usepackage{amsthm}%
\usepackage{mathrsfs}%
\usepackage[title]{appendix}%
\usepackage{xcolor}%
\usepackage{textcomp}%
\usepackage{manyfoot}%
\usepackage{booktabs}%
\usepackage[linesnumbered,ruled,vlined]{algorithm2e}
\usepackage{algpseudocode}%
\usepackage{listings}%
\usepackage{multirow}
\usepackage{colortbl}

\usepackage{booktabs}

\usepackage{enumitem} 

\theoremstyle{thmstyleone}%
\theoremstyle{thmstyletwo}%

\theoremstyle{thmstylethree}%

\begin{document}

\title[Article Title]{A data- and compute-efficient chest X-ray foundation model beyond aggressive scaling}
% \title[Article Title]{CheXficient: A data- and compute-efficient strategy for building high-performing CXR foundation models beyond aggressive scaling}


\author*[1,2]{\fnm{Chong} \sur{Wang}}
\email{chongwa@stanford.edu}
\equalcont{These authors contributed equally to this work.}

\author[1,2]{\fnm{Yabin} \sur{Zhang}}
% \email{yabin@stanford.edu}
\equalcont{These authors contributed equally to this work.}

\author[1,2]{\fnm{Yunhe} \sur{Gao}}
% \email{iiiauthor@gmail.com}

\author[1,2,3]{\fnm{Maya} \sur{Varma}}

\author[1,2]{\fnm{Clemence} \sur{Mottez}}

\author[1,2]{\fnm{Faidra} \sur{Patsatzi}}

\author[1,2]{\fnm{Jiaming} \sur{Liu}}

\author[1,4]{\fnm{Jin} \sur{Long}}

\author[1,2]{\fnm{Jean-Benoit} \sur{Delbrouck}}

\author[2]{\fnm{Sergios} \sur{Gatidis}}

\author[1,2,5]{\fnm{Akshay S.} \sur{Chaudhari}}

\author[1,2,5,6]{\fnm{Curtis P.} \sur{Langlotz}}

\affil[1]{\orgdiv{Stanford Center for Artificial Intelligence in Medicine and Imaging}, \orgname{Stanford University}, \orgaddress{\city{Palo Alto}, \state{CA}, \country{USA}}}

\affil[2]{\orgdiv{Department of Radiology}, \orgname{Stanford University}, \orgaddress{\city{Stanford}, \state{CA}, \country{USA}}}

\affil[3]{\orgdiv{Department of Computer Science}, \orgname{Stanford University}, \orgaddress{\city{Stanford}, \state{CA}, \country{USA}}}

\affil[4]{\orgdiv{Department of Pediatrics}, \orgname{Stanford University}, \orgaddress{\city{Stanford}, \state{CA}, \country{USA}}}

\affil[5]{\orgdiv{Department of Biomedical Data Science}, \orgname{Stanford University}, \orgaddress{\city{Stanford}, \state{CA}, \country{USA}}}

\affil[6]{\orgdiv{Department of Medicine}, \orgname{Stanford University}, \orgaddress{\city{Stanford}, \state{CA}, \country{USA}}}

% \affil[6]{\orgdiv{Department}, \orgname{Organization}, \orgaddress{\street{Street}, \city{City}, \postcode{610101}, \state{State}, \country{Country}}}


\abstract{

Foundation models for medical imaging are typically pretrained on increasingly large datasets, following a “scale-at-all-costs” paradigm. 
However, this strategy faces two critical challenges: 
large-scale medical datasets often contain substantial redundancy and severe class imbalance that bias representation learning toward over-represented patterns, 
% and indiscriminate training on heterogeneous data incurs considerable computational inefficiency.
and indiscriminate training regardless of heterogeneity in data quality incurs considerable computational inefficiency.
Here we demonstrate active, principled data curation during pretraining can serve as a viable, cost-effective alternative to brute-force dataset enlargement. 
We introduce CheXficient, a chest X-ray (CXR) foundation model that selectively prioritizes informative training samples. 
CheXficient is pretrained on only 22.7\% of 1,235,004 paired CXR images and reports while consuming under 27.3\% of the total compute budget, yet achieving comparable or superior performance to its full-data counterpart and other large-scale pretrained models.
% across diverse evaluation protocols. 
% We comprehensively assess CheXficient across 20 individual benchmarks spanning 5 task types.
% Non-adapted (off-the-shelf) evaluations include zero-shot findings classification (47 thoracic findings) and cross-modal retrieval (image to report and report to image).
% Adapted downstream evaluations encompass disease prediction (21 diseases), semantic segmentation (anatomical structures and abnormalities), and radiology report generation. 
We assess CheXficient across 20 individual benchmarks spanning 5 task types, 
including non-adapted off-the-shelf evaluations (zero-shot findings classification and cross-modal retrieval) and adapted downstream tasks (disease prediction, semantic segmentation, and radiology report generation). 
% Finally, we conduct a series of analysis to identify the data prioritization characteristic of CheXficient.
Further analyses show CheXficient systematically prioritizes under-represented training samples, improving generalizability on long-tailed or rare conditions. 
Overall, our work offers practical insights into the data and computation demands for efficient pretraining and downstream adaptation of medical vision–language foundation models. 
% Together, these results suggest that intelligent data curation can substantially reduce the data and computational demands of medical vision–language foundation models, providing a scalable pathway toward more efficient and deployable medical AI systems.

% In particular, we observe that CheXficient’s downstream performance scales well with the quantity and diversity of training data, demonstrating that image-only supervision is a scalable approach for training a foundational biomedical image encoder. 
% Notably, we observe that downstream performance scales favorably with both the quantity and diversity of curated training data, illustrating that image-only supervision is a scalable and efficient approach for training biomedical foundation models.

}




\maketitle

\section{Introduction}\label{sec1}

% \begin{figure}[!t]
% \centering
% \includegraphics[width=1.0\linewidth]{figures/medical_foundation_models.png}
% \vspace{-20 pt}
% \caption{Training data scale used in recent medical foundation models. \chong{is the figure needed?} \chong{size of circle represents data volume, also show performance?}
% }
% \label{fig:data_size}
% % \vspace{-10pt}
% \end{figure}

% Foundation models~\cite{bommasani2021opportunities} have emerged as powerful general-purpose learners capable of acquiring broad, transferable representations from large-scale data. 
% Their transformative impact in natural language processing~\cite{brown2020language} and computer vision~\cite{awais2025foundation} has catalyzed growing interest in extending these methodologies to medical domains, spanning radiology~\cite{wiggins2022opportunities}, pathology~\cite{lu2024visual}, dermatology~\cite{yan2025multimodal}, and ophthalmology~\cite{zhou2023foundation}, with the goal of developing versatile medical artificial intelligence (AI) systems that can support a wide spectrum of downstream tasks. 
% These foundation models are typically pretrained on massive medical datasets to learn highly generalizable feature representations, which then can be adapted effectively to specific applications with comparatively limited task-specific labels~\cite{he2024foundation}. 

Medical foundation models have emerged as a powerful paradigm for building general-purpose artificial intelligence systems in healthcare, offering a unified framework for learning transferable representations from massive medical data~\cite{bommasani2021opportunities,moor2023foundation,khan2025comprehensive}. 
In medical imaging, such models have shown strong performance across diverse specialties, including radiology~\cite{wiggins2022opportunities}, pathology~\cite{lu2024visual}, dermatology~\cite{yan2025multimodal}, and ophthalmology~\cite{zhou2023foundation}, with the potential to support a wide range of clinically relevant downstream tasks. 
Typically, medical foundation models are pretrained on indiscriminately collected, large-scale datasets (e.g., chest X-rays and associated free-text radiology reports~\cite{tiu2022expert}, figures with captions from scientific articles~\cite{zhang2023biomedclip,lozano2025biomedica}, and social media posts~\cite{huang2023visual}) to acquire highly generalizable representations, 
which can then be adapted to specific clinical applications using comparatively limited task-specific annotations~\cite{he2024foundation}.

Despite these promising achievements, the reliance on large-scale medical datasets presents significant practical challenges that limit the feasibility and sustainability of medical foundation models in real-world clinical settings. 
% One of the critical challenges is how to efficiently leverage the growing availability of large-scale medical pretraining data. 
In practice, retrospective archives, data-sharing initiatives, and web-sourced medical content have made large-scale medical datasets increasingly accessible~\cite{acosta2022multimodal}, 
yet standard pretraining approaches that blindly train on all available data face two critical inefficiencies. 
% While retrospective archives and data-sharing initiatives have made large-scale medical datasets increasingly available, 
% standard pretraining approaches that uniformly train on all available data face two critical inefficiencies. 
First, medical datasets often exhibit high redundancy and severe class imbalance~\cite{zhou2021review}, e.g., normal findings and common pathologies dominate while rare but clinically important conditions are under-represented, biasing representation learning toward over-represented patterns. 
Second, exhaustive training on millions of samples without accounting for heterogeneity in data quality incurs prohibitive computational burden~\cite{moor2023foundation} (e.g., RadFM~\cite{wu2025towards} trained on 16M radiology scans using 32 NVIDIA A100 GPUs for weeks; MedGemma~\cite{sellergren2025medgemma} trained on 33M medical image-text pairs using industrial-scale computational resources accumulating thousands of TPU-hours), 
creating substantial barriers for academic and clinical institutions with limited resources.

% For example, simply scaling up data volume is computationally prohibitive and often yields diminishing returns due to the high redundancy and long-tailed data distribution inherent in medical datasets.

% While recent work has explored synthetic data generation~\cite{bluethgen2025vision,sun2025data} to complement training samples, such approaches introduce additional computational overhead and raise concerns about fidelity and clinical validity. 
% To address data inefficiency, prior studies~\cite{bluethgen2025vision,sun2025data} have explored generative AI techniques~\cite{rombach2022high} to synthesize visually plausible medical training samples. However, such approaches are inherently constrained by the fidelity and clinical reliability of generated data, while simultaneously incurring additional computational costs due to the inclusion of large volumes of synthetic samples. 
% In contrast,
% Rather than pursuing ever-larger datasets,
% Addressing these inefficiencies requires a fundamental shift from the prevailing ``more data is better" paradigm.
% These challenges motivate a critical question: must medical foundation models be trained on all available data, or can strategic data utilization achieve comparable performance more efficiently?
These challenges motivate exploring an alternative paradigm, rather than pursuing ever-larger datasets.
Recent advances in machine learning suggest that optimizing which data are utilized for training can be more influential than simply increasing how much data is used~\cite{joshi2024data}. 
A growing body of work~\cite{gadre2023datacomp,cherti2023reproducible,mindermann2022prioritized,joshi2024data} shows that training samples differ markedly in their contribution to representation learning: some examples are highly informative, while others provide little benefit or can even hinder optimization~\cite{sorscher2022beyond}. 
Nevertheless, standard pretraining approaches of foundation models typically treat all data points as equally valuable, overlooking substantial differences in informativeness, representativeness, relevance, or diversity. 
This insight has driven a shift toward data curation or selection strategies, in which compact, high-value subsets are selectively emphasized and employed for training. 
Empirical evidence~\cite{gadre2023datacomp,wang2023too,sorscher2022beyond,goyal2024scaling} outside the medical domain demonstrates that models trained on carefully curated datasets can match or even surpass the performance of those trained on much larger corpora, challenging the long-standing assumption that generalization ability scales predominantly with dataset size. 
% , suggesting the potential of such strategies beyond dataset scale alone.
% Together, these findings suggest that how training data are selected and prioritized should be a fundamental consideration in designing the pretraining approaches.
These findings are particularly relevant to medical imaging, where data distributions are highly imbalanced across normal findings, rare pathologies, and acquisition protocols, treating all training samples uniformly is especially inefficient and can bias representation learning toward over-represented patterns. 
These observations raise a central question: 
can medical foundation models achieve comparable or superior performance while substantially reducing data and computational requirements through systematic, principled data selection?

% strategic 

\begin{figure}[!t]
\centering
\includegraphics[width=1.0\linewidth]{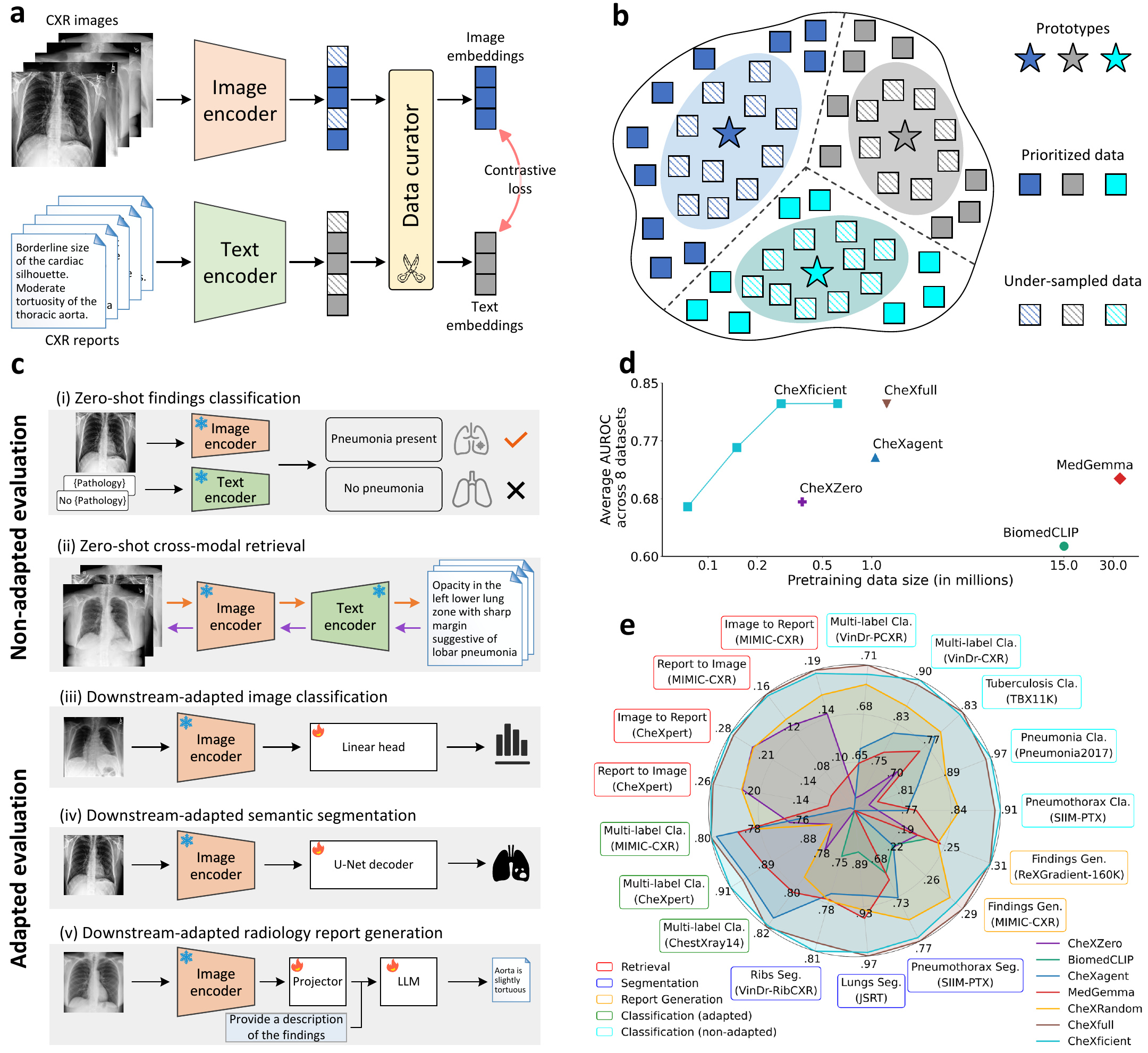}
\vspace{-10 pt}
\caption{Overview of this study. 
(a) \textbf{CheXficient pretraining strategy.} Paired chest X-ray (CXR) images and radiology reports are employed. 
A data curator selectively determines which image–report pairs are included in training, followed by optimization with an InfoNCE~\cite{oord2018representation} contrastive loss.
(b) \textbf{Data curation mechanism.} The data curator prioritizes image–report pairs based on their proximity to a set of learnable prototypes. Samples farther from the prototypes are assigned higher priority, while more redundant data closer to the prototypes is down-sampled.
(c) \textbf{Evaluation protocol.} After pretraining, CheXficient is evaluated on non-adapted tasks without any architectural or weight modifications, including (i) zero-shot finding classification and (ii) zero-shot cross-modal retrieval. CheXficient can also be adapted to downstream tasks via fine-tuning, including (iii) multi-class disease prediction, (iv) semantic segmentation, and (v) radiology report generation.
(d) \textbf{Data efficiency analysis.} Zero-shot findings classification performance (average AUROC on 8 datasets) versus pretraining data size compared with alternative pretraining strategies and models. 
(e) \textbf{Overall performance.} CheXficient achieves better or comparable performances with full-data pretraining, while greatly outperforming other existing large-scale pretrained models on diverse evaluation benchmarks spanning 5 task types: retrieval (in Recall@1), segmentation (in Dice score), and report generation (in RadGraph), adapted and non-adapted classification (in AUROC). 
% \chong{add pretraining data volume for each model name?}
}
\label{fig:workflow}
% \vspace{-10pt}
\end{figure}

% [
% To explore data-efficient strategies for building medical foundation models, chest X-ray (CXR) imaging is an appealing testbed for several reasons. 
% First, over 1.4 billion CXRs are performed annually worldwide, given their low cost and crucial role as an initial diagnostic tool; this abundance provides a unique opportunity for improving CXR interpretation and reporting workflows. 
% Second, many CXR foundation models, such as CheXagent~\cite{chen2024chexagent}, depend heavily on aggressive data expansion using millions of real-world images, making the domain a strong benchmark for evaluating the effectiveness of data-efficient strategy for building medical foundation models. 
% Third, CXRs capture clinically informative features for a broad spectrum of thoracic diseases or findings (over 174 types~\cite{bustos2020padchest}), enabling comprehensive assessment of a model’s generalization capability.
% % and several studies have demonstrated that data curation are valuable and effective.
% ]

In this work, we introduce CheXficient, a chest X-ray (CXR) foundation model built within a contrastive language-image pretraining (CLIP~\cite{radford2021learning}) framework and unprecedentedly designed to improve both data and computation efficiency through systematic data curation. 
As illustrated in Figure~\ref{fig:workflow}(a), CheXficient incorporates a prototype-driven online data curator during pretraining. 
A set of evolving prototypes (i.e., prototypical centroids)~\cite{zhou2022rethinking,wang2025mixture} is leveraged to approximate the underlying data manifold, enabling dynamic prioritization of informative CXR image–report data pairs for model optimization. 
To be concrete, training samples that lie farther from the prototypes (corresponding to under-represented but informative regions of the data distribution) are emphasized, while training samples near the prototypes, which tend to contain redundant information, are down-weighted and under-sampled (Figure~\ref{fig:workflow}(b); see Methods). 
This strategy yields a compact yet comprehensive coverage of the data manifold, leading to improved data scalability and computational efficiency compared with existing vision–language pretraining approaches (Figure~\ref{fig:workflow}(d)). 
We systematically evaluate CheXficient under 5 complementary evaluation protocols (Figure~\ref{fig:workflow}(c)),
encompassing both non-adapted tasks (zero-shot findings classification and zero-shot cross-modal retrieval) and adapted downstream tasks (multi-class disease prediction, semantic segmentation of anatomical structures and abnormalities, as well as radiology report generation). 
Together, these tasks reflect core clinical use cases of medical foundation models, spanning diagnostic decision support, anatomical understanding, and clinical reporting. 
Using only 22.7\% of 1,235,004 paired CXR training samples (sourced from 13 public datasets across multiple countries (Table~\ref{tab:pretraining_data_statistics})) and under 27.3\% of the full compute budget, CheXficient obtains performance on par with or superior to its full-data pretraining counterpart across 20 evaluation benchmarks (Figure~\ref{fig:workflow}(d, e)). 
In addition, CheXficient consistently outperforms existing state-of-the-art vision-language foundation models pretrained on substantially larger-scale medical data (Figure~\ref{fig:workflow}(d, e)). 
% Together, these results reveal that active, principled prototype-driven data curation enables effective data- and compute-efficient pretraining of medical foundation models, with broad implications for scalable and sustainable learning from large-scale medical imaging data. 

% the full pretraining data is pretrained on over 2 million images sourced from 11 institutions across multiple countries, covering 4 imaging modalities spanning diverse dermatological conditions (Fig. 1a–c). 

\section{Results}\label{sec2}

\subsection{Model design and experimental setup}

To validate the data- and compute-efficiency of CheXficient in contrastive CXR pretraining,
we construct a large-scale pretraining database containing over 1.235 million CXR image–report pairs collected from 13 publicly available sources, 
including CheXpert-Plus~\cite{chambon2024chexpert}, MIMIC-CXR~\cite{johnson2019mimic}, ReXGradient-160K~\cite{zhang2025rexgradient}, PadChest~\cite{bustos2020padchest}, BIMCV COVID-19~\cite{vaya2020bimcv}, CANDID-PTX~\cite{feng2021curation}, CASIA-CXR~\cite{metmer2024open}, Open-I~\cite{demner2015preparing}, ChestX-ray14~\cite{wang2017chestx}, BRAX~\cite{reis2022brax}, VinDr-CXR~\cite{nguyen2022vindr}, VinDr-PCXR~\cite{pham2023pedicxr}, and ChestDR~\cite{wang2023real}.
% For each source dataset, we strictly follow its official train/validation/test split when available.
The detailed composition of the full pretraining database is provided in Table~\ref{tab:pretraining_data_statistics}.

We compare the performance of CheXficient against that of the following baseline foundation models: CheXrandom and CheXfull. 
To obtain a fair comparison, the same network architectures (DINOv2~\cite{oquab2023dinov2} as the image encoder and BioClinicalBERT~\cite{alsentzer2019publicly} as the text encoder), pretraining recipe (image-text contrastive learning by InfoNCE loss~\cite{oord2018representation}), and evaluation protocols are employed in all models. 
CheXfull is pretrained on the entire 1.235 million paired CXR data. 
CheXficient is pretrained on a carefully curated subset from the full pretraining corpus.  
CheXrandom is pretrained on a random subset of the full set, with an identical size as the curated subset. 
We evaluate CheXficient on 20 benchmarks across two evaluation protocols: (1) non-adapted zero-shot evaluation: findings classification (47 thoracic findings) and cross-modal retrieval (image to reports, image to \emph{Findings}, image to \emph{Impression} and vice versa); (2) adapted downstream fine-tuning evaluation: multi-disease prediction, radiology report generation, as well as semantic segmentation of anatomic structures and abnormalities. 
Detailed information on downstream datasets is given in Table~\ref{tab:evaluation_data_statistics}. 
The area under the receiver operating curve (AUROC), Recall@1, Dice score, and standard radiology report generation metrics (e.g., RadGraph) are utilized for evaluating the task performance of these models. 
In experiments, we train the models with different seeds to obtain multiple results and evaluate their performances for statistical analysis. 
% We assess statistical significance by performing two-sided $t$-tests and calculating the $p$-values between CheXficient and CheXfull for each task, with $p<$ 0.05 indicating a statistically significant difference.

\subsection{CheXficient prioritizes under-represented training data}

\begin{figure}[!t] 
\centering
\includegraphics[width=1.0\linewidth]{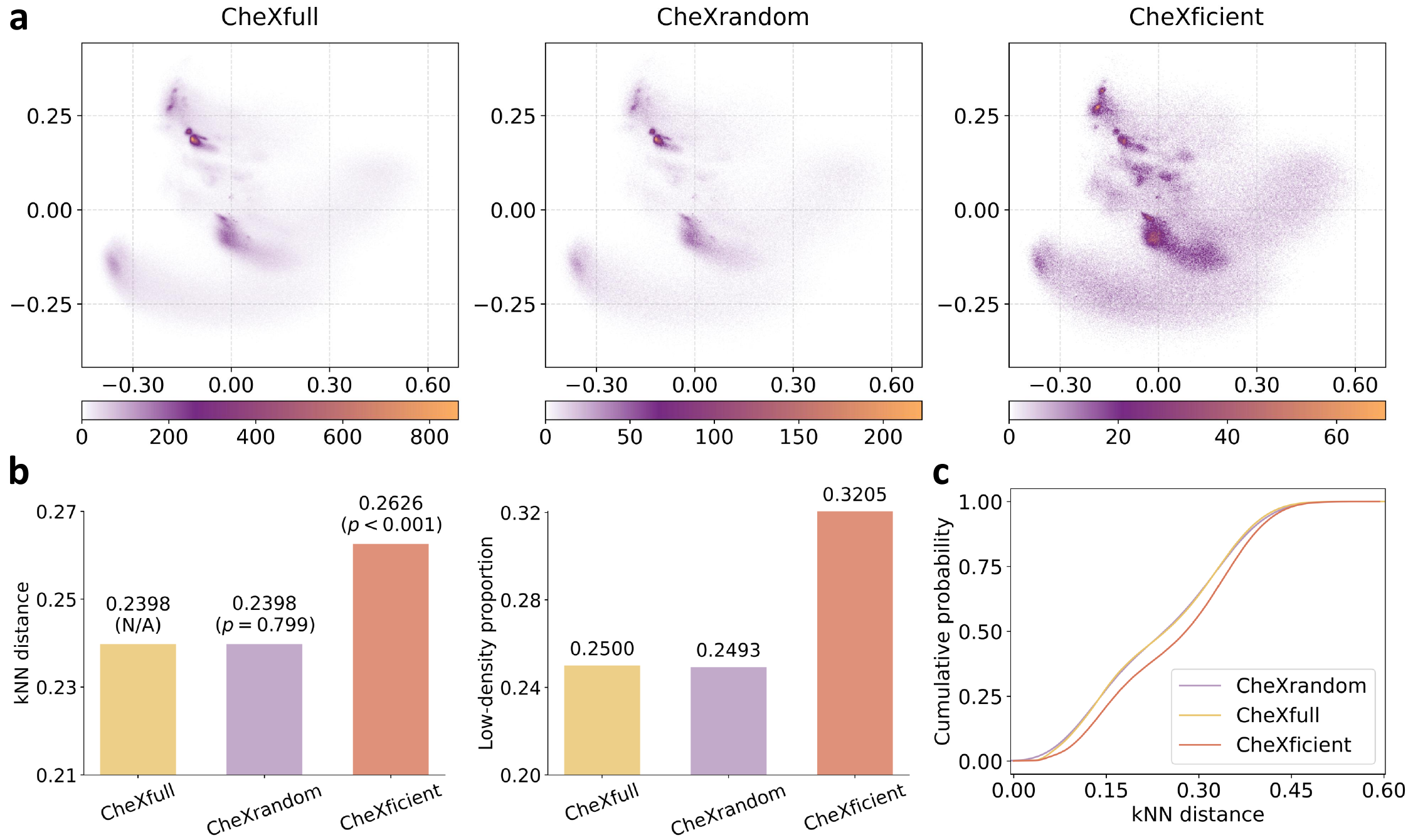}
\vspace{-15 pt}
\caption{
(a) Feature distribution of training samples from the full set (CheXfull, $n$ = 1,235K), random subset (CheXrandom, $n$ = 280K), and curated subset (CheXficient, $n$ = 280K).
% Two-dimensional PCA visualization of feature embeddings using histograms.
Qualitative histograms of feature embeddings projected onto a two-dimensional PCA space (Colorbars are independently normalized to reveal the structural organization of the PCA space, absolute density values are therefore not directly comparable across methods).
% Colorbars are shared between CheXrandom and CheXficient to enable direct density comparison.
The curated subset occupies distinct and long-tailed regions in the feature space, compared with both the full and random subsets.
(b) Quantitative analysis of local feature density (left) for samples from the full set, random subset, and curated subset, 
measured by the average $k$-nearest neighbor (kNN) distance ($k=20$) computed in the raw feature space.
$p$-values are reported from Welch’s t-test by comparing each subset against the full set. 
The low-density proportion (right) denotes the fraction of samples falling within the top 25\% lowest-density regions of the full dataset. 
(c) CDF plot of the average kNN distance.
Curated samples (red) are shifted toward higher kNN distances, indicating enrichment in low-density regions,
whereas the random subset (purple) closely overlaps with the full set (yellow). 
}
\label{fig:feat_distribution}
% \vspace{-10pt}
\end{figure}

% \begin{table*}[h]
% \centering
% \setlength{\extrarowheight}{0pt}
% \addtolength{\extrarowheight}{\aboverulesep}
% \addtolength{\extrarowheight}{\belowrulesep}
% \setlength{\aboverulesep}{0pt}
% \setlength{\belowrulesep}{0pt}
% \caption{
% Quantitative analysis of feature density for the full training set (CheXfull), random subset (CheXrandom), and curated subset (CheXficient).
% Local sample density is measured by the mean $k$-nearest neighbor (kNN) distance ($k=20$) computed in the raw feature space.
% The low-density ratio denotes the proportion of samples falling within the top 25\% lowest-density regions of the full dataset. 
% $p$-values are reported from two-sided Welch’s t-test comparing each subset against the full set. \chong{move to Appendix.}
% }
% \vspace{-6pt}
% \label{tab:feat_kNN}

% \setlength{\tabcolsep}{4.8 mm}
% \resizebox{0.99999\linewidth}{!}{

% \begin{tabular}{lccc} 
% \hline
% Method                            & CheXfull           & CheXrandom                 & CheXficient                \\ 
% \hline
% Mean kNN distance ($p$-value)     & 0.2398 (N/A)       & 0.2398 ($p = 0.368$)       & 0.2626 ($p < 0.001$)       \\
% Low-density ratio                 & 0.2500             & 0.2493                     & 0.3205                     \\
% \hline
% \end{tabular}

% }
% \end{table*}

To study and understand the characteristics of the curated training data, we extract feature embeddings for full-set and curated-set training samples, using the pretrained CheXfull model.
For comparison, we additionally consider a random subset (with the same size as the curated set) used by CheXrandom. 
We apply principal component analysis~\cite{abdi2010principal} (PCA) to project the high-dimensional features (unified multimodal representation, see Section~\ref{sec:algorithm}) into a two-dimensional space for visualization.
Figure~\ref{fig:feat_distribution}(a) illustrates the resulting feature distributions using two-dimensional histograms.
The curated subset exhibits a substantially different pattern compared to both the full dataset and the random subset.
Particularly, curated samples populate regions that are sparsely covered by the full dataset, suggesting an over-representation of uncommon imaging patterns, atypical report semantics, or long-tailed thoracic diseases.

To quantitatively characterize these differences, we also measure the local sample density using the average $k$-nearest neighbor (kNN) distance in the raw feature space.
As reported in Figure~\ref{fig:feat_distribution}(b), curated samples exhibit a much larger kNN distance than the full set (0.2626 vs. 0.2398), indicating an enrichment of long-tailed and under-represented samples in low-density regions.
This difference is statistically significant ($p < 0.001$).
In addition, more than 32\% of curated samples fall within the top 25\% lowest-density regions of the full set. 
In Figure~\ref{fig:feat_distribution}(c), the CDF (cumulative distribution function) plot of average kNN distances shows that the curated subset (red) is systematically shifted to the right, further verifying an enrichment of low-density, long-tailed samples. 
In contrast, a randomly sampled subset (purple) closely overlaps with the full set (yellow), demonstrating that random selection does not preferentially sample low-density regions.

% \chong{show some pruned examples?}. 

% 1. Fréchet Distance / FID
% 2. Maximum Mean Discrepancy (MMD)

% 3. To quantitatively assess the representation differences between the curated and full training sets beyond visual inspection, we analyze feature separability on long-tailed disease categories.
% Tail findings are defined as disease categories whose prevalence ranks in the bottom 20\% of the training set.
% For these tail classes, we compute the \textbf{Fisher Discriminant Ratio (FDR)}, measuring the ratio between inter-class distances and intra-class variances in the feature space.
% CheXficient consistently achieves higher FDR values than CheXfull, indicating improved inter-class separability and reduced intra-class dispersion for rare diseases.
% We further observe higher \textbf{kNN purity on tail classes}, confirming that curated pretraining emphasizes more discriminative representations for under-represented findings.

\subsection{Performance in non-adapted zero-shot tasks}
\label{sec:zeroshot_evaluation}

We first assess the performance of pretrained vision–language models (CheXficient, CheXrandom, and CheXfull) on two non-adapted zero-shot tasks, where these models are directly used without any architecture or weight modifications.
% We prompt the text encoder with (1) class-specific prompts~\cite{tiu2022expert} for classification (Section ) or (2) radiology reports for cross-modal retrieval.
For zero-shot findings classification, we follow prior work~\cite{tiu2022expert} and prompt the text encoder with class-specific descriptions. 
For zero-shot cross-modal retrieval, we employ radiology reports as queries to prompt the text encoder.

\subsubsection{Zero-shot findings classification}

\begin{figure}[!t] 
\centering
\includegraphics[width=1.0\linewidth]{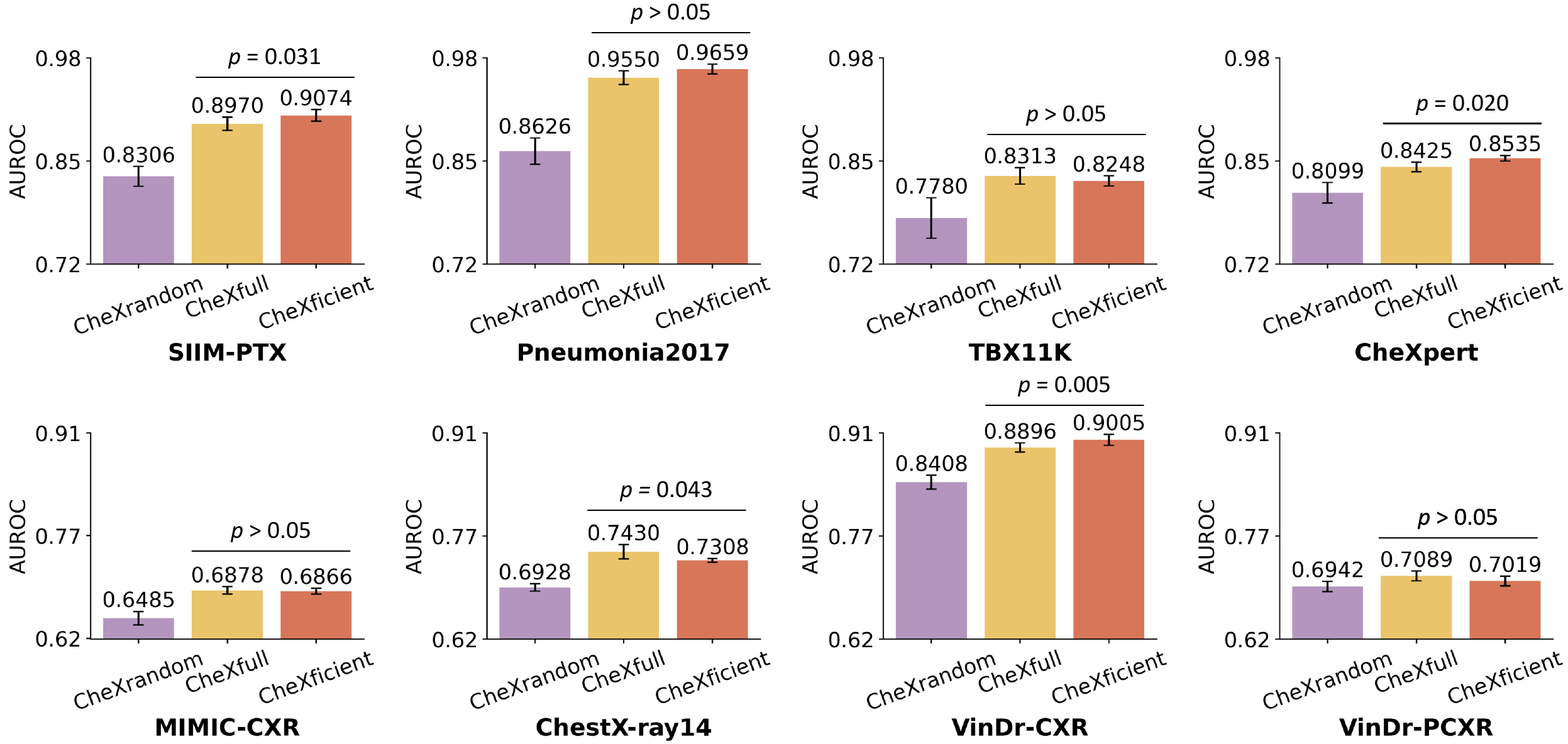}
\vspace{-15 pt}
\caption{Performance on zero-shot findings classification. 
We evaluate the pretrained models on 8 public datasets. 
Among them, SIIM-PTX, Pneumonia2017, and TBX11K are from external domains unseen in pretraining, while the remaining 5 datasets are used for internal evaluation.
Compared to CheXfull, CheXficient achieves higher AUROC on three datasets ($p < 0.05$), comparable performance on four datasets ($p > 0.05$), and lower performance on one dataset ($p < 0.05$).
Across all datasets, CheXficient achieves higher AUROC than CheXrandom.
We present the mean $\pm$ 95\% confidence interval (CI) of AUROC for each model. The listed $p$-values are computed using two-sided $t$-tests. 
}
\label{fig:zero-shot classification}
% \vspace{-10pt}
\end{figure}

% Pneumothorax (PTX) occurs when air enters the pleural space, potentially causing partial or complete lung collapse and requiring urgent clinical management.
We evaluate the pretrained CheXficient, CheXrandom, and CheXfull on the SIIM-PTX~\cite{SIIM-ACR-Pneumothorax-Segmentation-2019} dataset ($n$ = 1,372), containing binary classification labels (pneumothorax vs. no pneumothorax). 
As shown in Figure~\ref{fig:zero-shot classification}, CheXficient significantly outperforms CheXfull ($p = 0.031$) and substantially exceeds the performance of CheXrandom (AUROC: 0.9074 (95\% CI 0.9001, 0.9148) vs. 0.8306 (95\% CI 0.8181, 0.8431)). 
% Pneumonia is an infection of the lung parenchyma that causes inflammation and consolidation of the air sacs, leading to symptoms like cough, fever, chills, and shortness of breath.
We also evaluate the above three pretrained models on the Pneumonia2017~\cite{kermany2018labeled} dataset ($n$ = 624), where each CXR image is labeled either pneumonia (bacterial/viral) or normal. 
CheXficient attains AUROC performance comparable to CheXfull, with an improvement of more than 10.3\% over CheXrandom.
% Tuberculosis is a chronic infectious disease primarily affecting the lungs and remains a global public health concern.
We conduct tuberculosis classification on the TBX11K~\cite{liu2020rethinking} dataset ($n$ = 1,800), where each image is labeled either tuberculosis or non-tuberculosis.
CheXficient again achieves AUROC performance comparable to CheXfull and outperforms CheXrandom.

Beyond single-disease tasks, recent research~\cite{ma2025fully,wang2025cross} has focused on comprehensive detection of multiple thoracic diseases and imaging findings.
We therefore evaluate CheXficient on five multi-label chest X-ray benchmarks: CheXpert~\cite{chambon2024chexpert} ($n$ = 500, 14 categories), MIMIC-CXR~\cite{johnson2019mimic} ($n$ = 3,082, 14 categories), ChestX-ray14~\cite{wang2017chestx} ($n$ = 25,596, 15 categories), VinDr-CXR~\cite{nguyen2022vindr} ($n$ = 3,000, 28 categories), and VinDr-PCXR~\cite{pham2023pedicxr} ($n$ = 1,397, 15 categories) datasets. 
For each dataset, we compute category-wise AUROC and report the macro-average. 
CheXficient achieves a higher macro AUROC than CheXfull on CheXpert  ($p =$ 0.020) and VinDr-CXR ($p =$ 0.005), while yielding comparable performance on MIMIC-CXR and VinDr-PCXR, and lower performance on ChestX-ray14.
Across all five benchmarks, CheXficient consistently achieves higher AUROC than CheXrandom.
% demonstrating strong generalization capacity in multi-disease identification.
% Detailed per-category AUROC results are provided in Extended Data~XXX. 

Notably, three datasets (SIIM-PTX, Pneumonia2017, and TBX11K) originate from institutions and data distributions entirely disjoint from the pretraining corpus and are therefore treated as external evaluation benchmarks (Table~\ref{tab:evaluation_data_statistics}). 
As shown in Figure~\ref{fig:zero-shot classification}, CheXficient achieves performance comparable to or exceeding CheXfull across these unseen domains, despite being pretrained on less data.
% These results underscore CheXficient’s generalization capability to unseen domains and highlight its potential for robust real-world deployment across diverse clinical settings.
AUPRC results on the 8 public datasets are reported in Extended Data Figure~\ref{fig:zero-shot classification AUPRC}.

% \begin{table*}[h]
% \centering
% \caption{Zero-shot classification. The results are averaged over three runs by default, with standard deviations shown in parentheses. The best results are highlighted in bold.
% % and the second best results are in underlined.
% }
% \vspace{-6pt}
% \label{tab:zeroshot_classification}

% \setlength{\tabcolsep}{1.0 mm}
% \resizebox{1.0\linewidth}{!}{

% \begin{tabular}{lcccccccc} 
% \hline
% Method                     & CheXpert                   & MIMIC-CXR                  & ChestX-ray14               & VinDr-CXR                  & VinDr-PCXR                 & TBX11K                     & SIIM-PTX                   & Pneumonia2017                    \\ 
% \hline
% CheXrandom                 & 0.8099 [0.7969, 0.8229]    & 0.6485 [0.6389, 0.6580]    & 0.6928 [0.6877, 0.6979]    & 0.8408 [0.8311, 0.8506]    & 0.6942 [0.6872, 0.7013]    & 0.7780 [0.7524, 0.8035]    & 0.8306 [0.8181, 0.8431]    & 0.8626 [0.8459, 0.8792]           \\
% CheXfull                   & 0.8425 [0.8364, 0.8485]    & 0.6878 [0.6824, 0.6931]    & 0.7430 [0.7329, 0.7530]    & 0.8896 [0.8831, 0.8960]    & 0.7089 [0.7020, 0.7158]    & 0.8313 [0.8209, 0.8417]    & 0.8970 [0.8887, 0.9052]    & 0.9550 [0.9464, 0.9636]           \\
% CheXficient                & 0.8535 [0.8501, 0.8569]    & 0.6866 [0.6824, 0.6908]    & 0.7308 [0.7281, 0.7336]    & 0.9005 [0.8929, 0.9081]    & 0.7019 [0.6951, 0.7087]    & 0.8248 [0.8183, 0.8313]    & 0.9074 [0.9001, 0.9148]    & 0.9659 [0.9597, 0.9722]           \\
% \hline
% \end{tabular}

% }
% \end{table*}

\subsubsection{Zero-shot cross-modal retrieval}

\begin{figure}[!t] 
\centering
\includegraphics[width=1.0\linewidth]{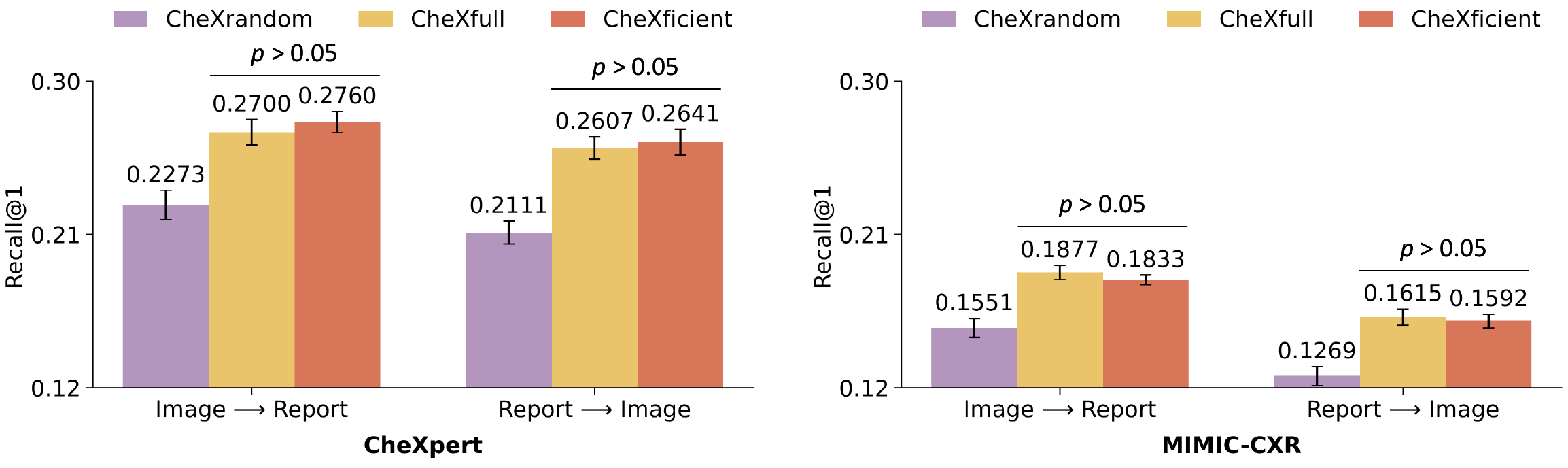}
\vspace{-15 pt}
\caption{Performance on zero-shot cross-modal retrieval.
We evaluate the pretrained models on the public CheXpert and MIMIC-CXR benchmarks. 
CheXficient achieves performance comparable to CheXfull on both datasets ($p > 0.05$), while outperforming CheXrandom.
We report the mean $\pm$ 95\% CI of Recall@1 (The recall of retrieving the exact paired CXR report (or image) within the top-1 result for a given CXR image (or report)). 
% The listed $p$ values are calculated using two-sided t-test.
% \chong{show Findings and Impressions instead?}
}
\label{fig:zero-shot retrieval}
% \vspace{-10pt}
\end{figure}

Zero-shot cross-modal retrieval aims to evaluate a model’s ability to match a CXR image to its corresponding radiology report (\emph{Findings} section, \emph{Impressions} section, or their combination) and vice versa. 
As illustrated in Figure~\ref{fig:zero-shot retrieval}, CheXficient obtains performances on par with CheXfull and surpasses CheXrandom, on the task of retrieving the correct CXR report for a given CXR image (Image $\rightarrow$ Report) across both evaluation datasets (CheXpert ($n$ = 1,000) and MIMIC-CXR ($n$ = 3,082)). 
A similar trend is observed for retrieving the correct CXR image given a CXR report (Report $\rightarrow$ Image). 
% These results suggest that CheXficient successfully selects valuable samples during pretraining, enabling the model to learn clinically meaningful features and to maintain strong alignment between visual and textual representations.
In this evaluation, \emph{Findings} and \emph{Impressions} sections are combined to form the textual report, retrieval results based solely on either section are provided in Extended Data Figure~\ref{fig:zero-shot retrieval Findings} and Figure~\ref{fig:zero-shot retrieval Impression}.

\subsubsection{Pretraining data and compute efficiency}

\begin{figure}[!t] 
\centering
\includegraphics[width=1.0\linewidth]{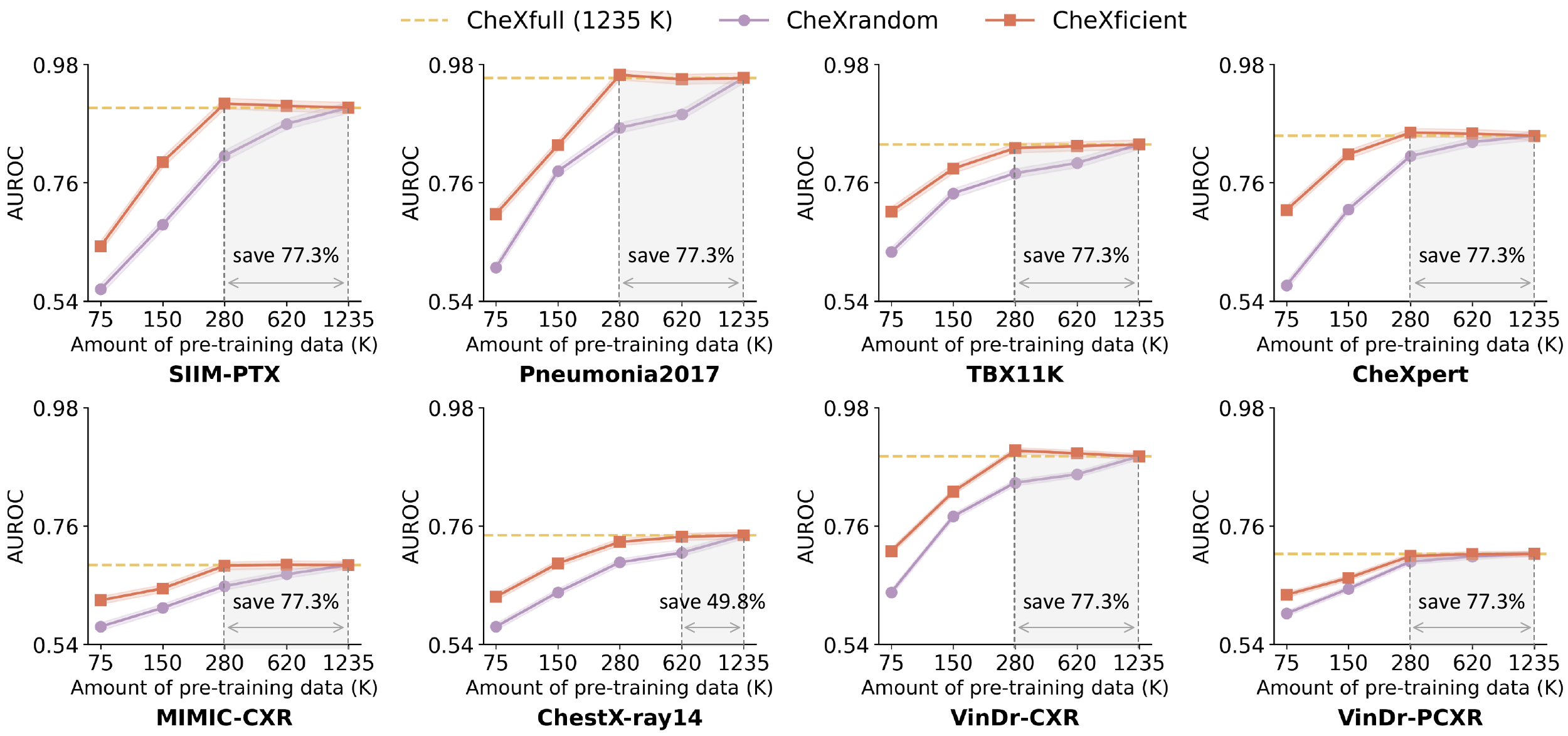}
\vspace{-15 pt}
\caption{
Zero-shot findings classification performance under different pretraining data budgets. 
The performance of both CheXficient and CheXrandom progressively increases on all 8 benchmarks as the quantity of pretraining samples increases. 
CheXficient consistently outperforms CheXrandom at all data budgets and across all benchmarks, demonstrating higher data efficiency.
When pretrained on only 280K curated image–report pairs (22.7\% of the full data, except for ChestX-ray14), CheXficient performs better or similarly to the full-data pretrained model CheXfull. 
}
\label{fig:pretraining data efficiency}
% \vspace{-10pt}
\end{figure}

We validate the importance and advantage of active curation by assessing the pretraining data and compute efficiency, defined as the amount of paired image–report data and pretraining time (GPU-hours) required to build a high-performing CXR foundation model.

We compare CheXficient with the baseline CheXrandom model, which is trained under identical pretraining data budgets ranging from 75K to 1.235M image–report pairs.
Figure~\ref{fig:pretraining data efficiency} shows zero-shot findings classification performance across different pretraining set sizes, with the CheXfull model (trained on the entire dataset) included as a reference.
The performance of both models consistently improves across the eight evaluation benchmarks as the size of the pretraining corpus increases.
Across all data budgets and benchmarks, CheXficient achieves higher performance than CheXrandom.
When pretrained on 280K curated image–report pairs (22.7\% of the full data), CheXficient achieves performance comparable to CheXfull on SIIM-PTX, Pneumonia2017, CheXpert, MIMIC-CXR, and VinDr-CXR.
% indicating that carefully data-curated pretraining can yield sufficiently strong medical representations even under constrained data budgets. 

% The AUPRC results are shown in Extended Data Fig. XXX. 

\begin{figure}[!t]
\centering
\includegraphics[width=1.0\linewidth]{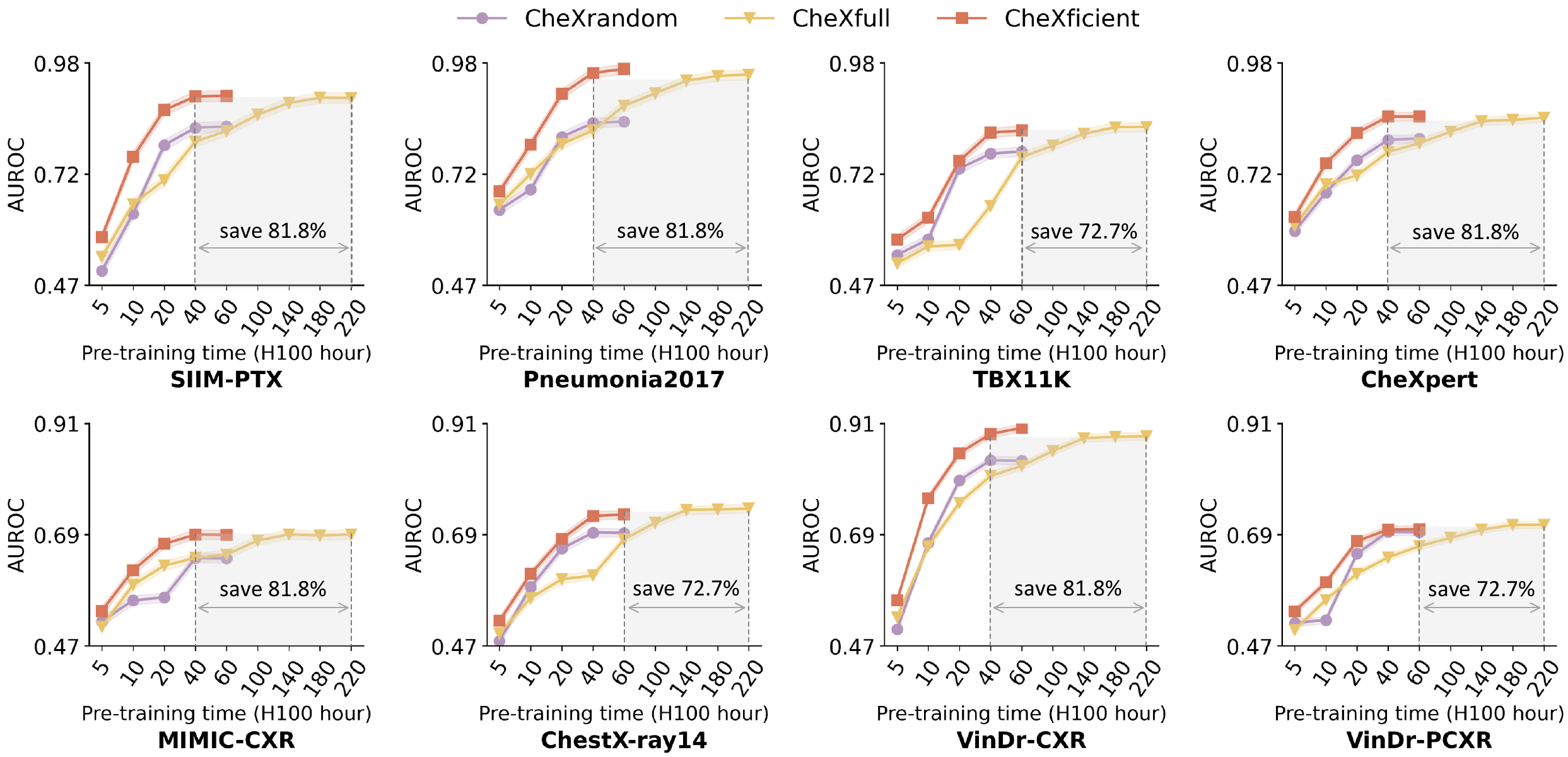}
\vspace{-15 pt}
\caption{
Zero-shot findings classification performance under different pretraining computation budgets. 
The three models are pretrained for the same number of epochs. 
CheXficient (280K) consistently outperforms CheXrandom (280K) and CheXfull (1235K) on all evaluation benchmarks when using the same pretraining time, highlighting the compute efficiency of curated pretraining of CheXficient.
}
\label{fig:pretraining time efficiency}
% \vspace{-10pt}
\end{figure}

We further investigate the pretraining compute efficiency in the context of zero-shot findings classification.
Computational cost is quantified in NVIDIA H100 GPU-hours.
For a fair comparison, all models are pretrained for the same total number of epochs (20 epochs), which we found sufficient for convergence.
CheXfull is pretrained on the entire dataset comprising 1.235 million image–report pairs, whereas CheXrandom and CheXficient are pretrained on the same subset size (280K).
As shown in Figure~\ref{fig:pretraining time efficiency}, CheXficient achieves comparable or superior performance while requiring substantially less pretraining time, saving 72.7–81.8\% of H100 GPU-hours relative to CheXfull, which requires 220 H100 GPU-hours. 
Under matched pretraining computational budgets, CheXficient consistently outperforms CheXrandom across all of the eight evaluation benchmarks.
% Together, these results demonstrate the advantage of data-curated pretraining in improving pretraining compute efficiency for chest X-ray foundation models.

% The AUPRC results are shown in Extended Data Fig. 4b. 
% We also show the pretraining loss (Supplementary Fig. 1). 

\subsection{Performance in adapted downstream tasks}

We next evaluate the performance of pretrained vision–language models (CheXficient, CheXrandom, and CheXfull) on three adapted downstream tasks: image classification, semantic segmentation, and radiology report generation.
In all tasks, the pretrained image encoder is leveraged as a fixed feature extractor, while task-specific heads or decoders are appended and fine-tuned on target labeled datasets. 
Detailed fine-tuning configurations are provided in Appendix~\ref{downstream details}.

\subsubsection{Classification label efficiency and fine-tuning performance}

To assess the representation quality learned through efficient pretraining in CheXficient, we evaluate downstream image classification (disease prediction) performance on three representative multi-label CXR datasets: CheXpert, MIMIC-CXR, and ChestXray14, which cover a diverse set of anatomical and pathological findings and diseases.
We adopt a linear-probing protocol with a frozen image encoder backbone. 
Specifically, a randomly initialized linear classification head is attached to the pretrained image encoder, and only this linear classifier is trained using images and corresponding labels from the target dataset.
For comparison purposes, we also include DINOv2~\cite{oquab2023dinov2}, which is pretrained solely on 142 million natural images.

\begin{figure}[!t]
\centering
\includegraphics[width=1.0\linewidth]{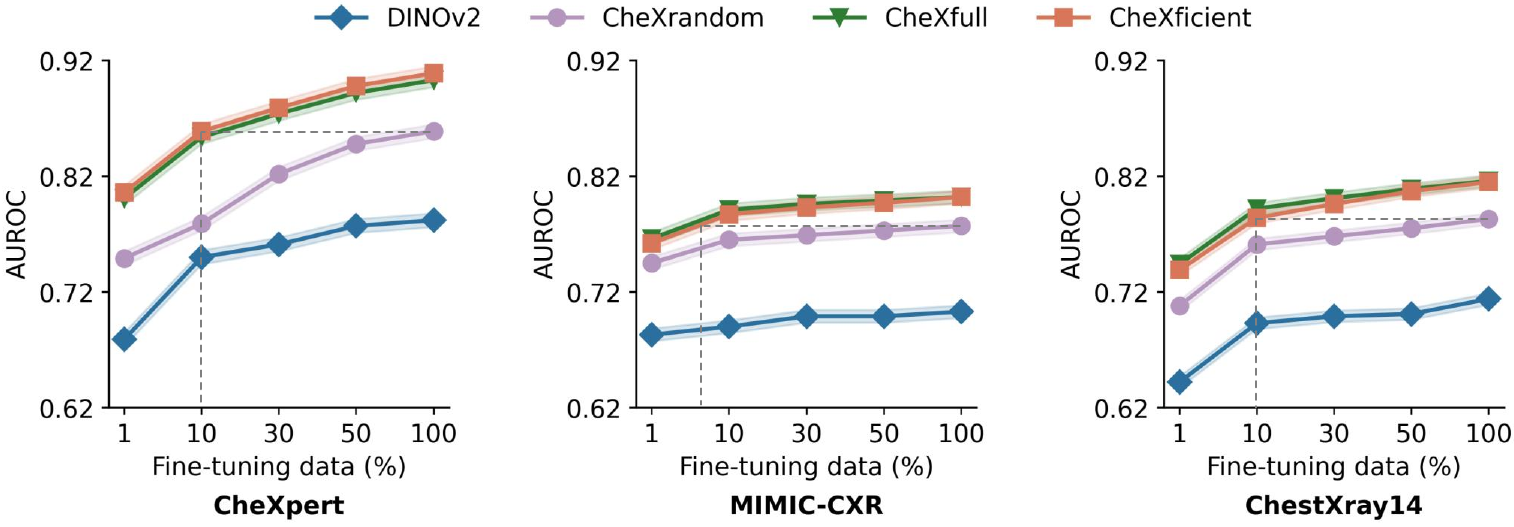}
\vspace{-15 pt}
\caption{
Comparison of CheXficient with other pretrained models in label-efficient disease prediction across three representative datasets: CheXpert ($n$ = 500, 14 categories), MIMIC-CXR ($n$ = 3,082, 14 categories), and ChestXray14 ($n$ = 25,596, 15 categories). 
Performance is reported at varying proportions of labeled training data. 
Vertical dashed lines indicate the amount of labeled data required for CheXficient to match the performance of its counterpart CheXrandom. 
}
\label{fig:classification label efficiency}
% \vspace{-10pt}
\end{figure}

As shown in Figure~\ref{fig:classification label efficiency}, AUROC performance of all models increases with the amount of labeled data used for fine-tuning.
Models pretrained on CXR data achieve higher performance than DINOv2.
% highlighting the advantage of domain-specific pretraining.
Across different proportions of labeled training data, CheXficient realizes performance comparable to CheXfull.
CheXficient achieves performance comparable to CheXrandom while using approximately 10\% of the labeled training samples.
The AUPRC results of CheXficient are also similar to those of CheXfull on these three datasets (Extended Data Figure~\ref{fig:classification label efficiency AUPRC}).
Furthermore, CheXficient also outperforms other large-scale pretrained models (e.g., BiomedCLIP~\cite{zhang2023biomedclip} and MedGemma~\cite{sellergren2025medgemma}) that are trained on more than 50$\times$ the amount of data used by CheXficient (see Figure~\ref{fig:workflow}(e) and Extended Data Table~\ref{tab:comparative overview}), underscoring the advantages of active and principled data curation over sheer dataset scaling.

\subsubsection{Segmentation annotation efficiency and fine-tuning performance}

\begin{figure}[!t] 
\centering
\includegraphics[width=1.0\linewidth]{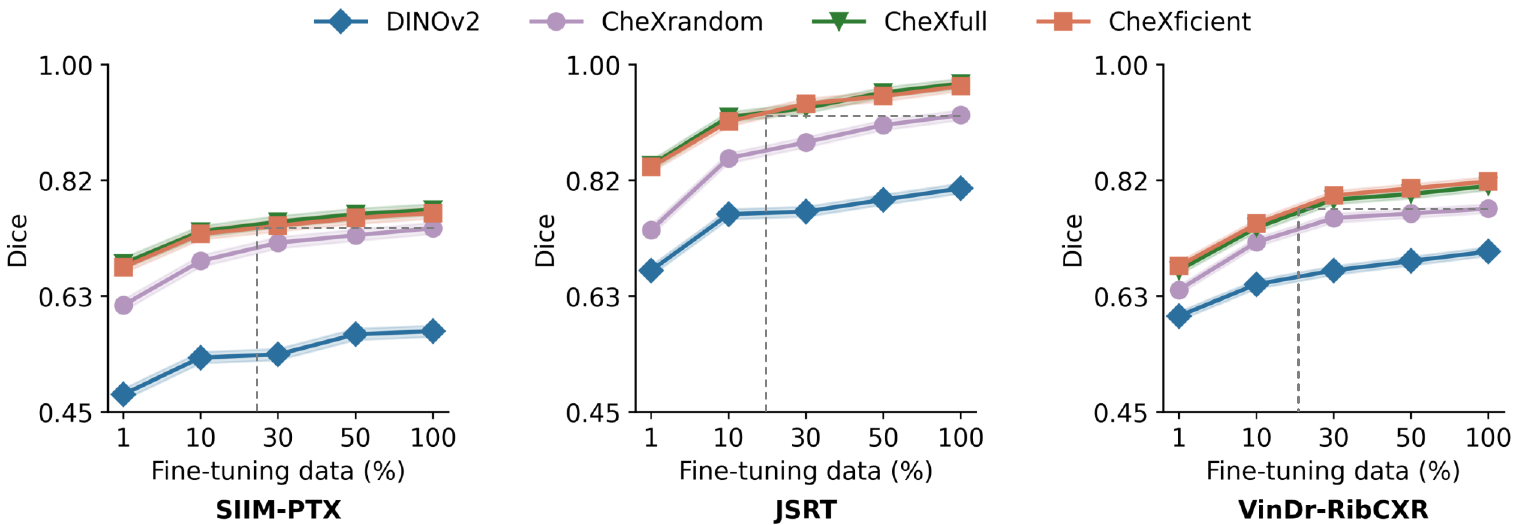}
\vspace{-15 pt}
\caption{
Comparison of CheXficient with other pretrained models in annotation-efficient anatomical and pathological segmentation across three datasets: SIIM-PTX ($n$ = 1,372), JSRT ($n$ = 50), and VinDr-RibCXR ($n$ = 49). 
Performance is reported at varying proportions of annotated training samples. 
Vertical dashed lines indicate the amount of annotated data required for CheXficient to match the performance of its counterpart CheXrandom.
}
\label{fig:segmentation annotation efficiency}
% \vspace{-10pt}
\end{figure}

% \textbf{Experimental setup.}
To further evaluate the quality of dense representations learned through curated pretraining, we focus on downstream tasks of semantic segmentation. 
Specifically, we consider three commonly adopted CXR benchmarks covering both pathological and anatomical structures: pneumothorax segmentation on SIIM-PTX~\cite{SIIM-ACR-Pneumothorax-Segmentation-2019}, lung segmentation on JSRT~\cite{shiraishi2000development}, and rib segmentation on VinDr-RibCXR~\cite{nguyen2021vindr}.
We employ a frozen image encoder backbone within an encoder–decoder segmentation framework. 
A U-Net~\cite{ronneberger2015u} decoder is attached to the pretrained image encoder and fine-tuned using CXR images and their corresponding pixel-wise annotations from each target dataset, while the image encoder remains fixed. 
This setup allows us to assess the quality and transferability of learned representations for pixel-level dense prediction tasks.
% and their high-level segmentation capability when coupled with a more expressive U-Net decoder.
% In addition, to contextualize the performance of our data-efficient pretraining approach, we also compare CheXficient against a range of existing pretrained models. 
% Note that all competing methods reported in Table~\ref{tab:downstream_segmentation} are fine-tuned using the same downstream training data to ensure a fair comparison.
% \chong{provide segmentation results on using linear decoder in Supplementary XXX.}

% \textbf{Result Analysis.}
As shown in Figure~\ref{fig:segmentation annotation efficiency}, CheXficient achieves segmentation performance comparable to CheXfull across all three tasks.
CheXficient exhibits strong annotation efficiency, matching the performance of CheXrandom while using 10–30\% of the densely-annotated training samples. 
These results demonstrate that curated, data-efficient pretraining enables the learning of high-quality image representations that transfer effectively to challenging downstream segmentation tasks.
CheXficient also achieves performance competitive with other off-the-shelf pretrained image encoders (e.g., BiomedCLIP~\cite{zhang2023biomedclip}, MedGemma~\cite{sellergren2025medgemma}, and CheXagent~\cite{chen2024chexagent}), in spite of being trained on substantially smaller-scale pretraining datasets (Figure~\ref{fig:workflow}(e)).

\subsubsection{Radiology report generation}

\begin{figure}[!t] 
\centering
\includegraphics[width=0.90\linewidth]{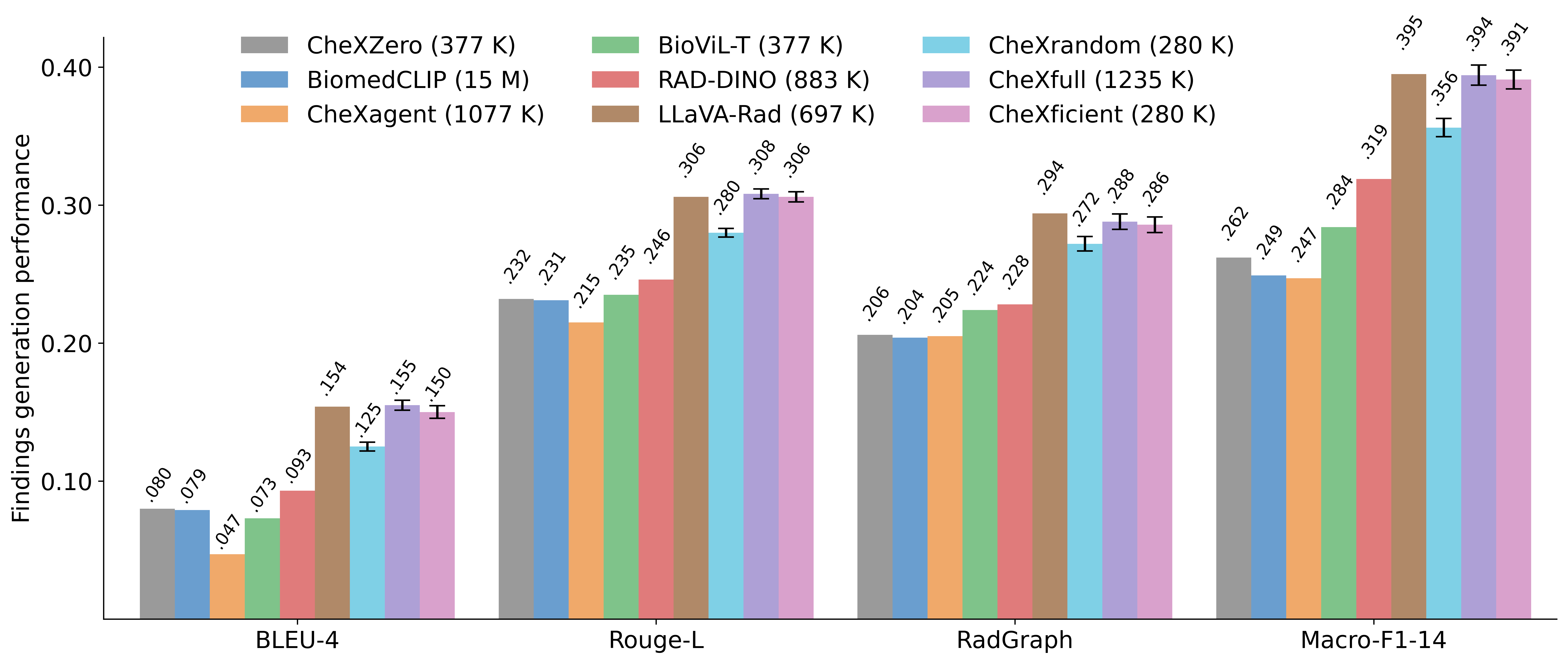}
\vspace{-5 pt}
\caption{
Downstream radiology report (\emph{Findings} section) generation performance (\%) on official test split of MIMIC-CXR ($n$ = 2,461). 
Text in parentheses denotes the pretraining data size of the model.
}
\label{fig:report generation MIMIC}
% \vspace{-10pt}
\end{figure}

% \textbf{Experimental setup.}
We also evaluate the quality of image representations learned through efficient pretraining on a vision–language downstream task, namely radiology report generation for the \emph{Findings} section.
% Following LLaVA-Rad~\cite{zambrano2025clinically} and RAD-DINO~\cite{perez2025exploring}, we conduct experiments on the MIMIC-CXR dataset using the official splits. 
% We exclude non-frontal images and samples lacking a \emph{Findings} section, resulting in a distribution of 146,909 training, 7,250 validation, and 2,461 testing image–text pairs.
% The training split is utilized for fine-tuning the language decoder. 
% An identical pre-processing pipeline is applied to ReXGradient-160K, yielding 140,000, 10,000, and 10,000 pairs for training, validation, and testing, respectively.
For the model architecture, we adopt a LLaVA-style multimodal framework~\cite{liu2023visual}. 
To generate the output report, patch embeddings extracted from the frozen image encoder are projected and concatenated with a textual instruction:
``Provide a description of the findings in the radiology image given the following Indication: $\langle \ \rangle$.'' 
If the Indication information is unavailable, the optional $\langle \ \rangle$ field is omitted. 
Consistent with LLaVA-Rad~\cite{zambrano2025clinically}, we employ a two-layer fully connected (MLP) projector and Vicuna-7B-v1.5~\cite{chiang2023vicuna} as the language model. 
The projector network is randomly initialized and jointly trained with the language decoder, while the image encoder remains frozen.
Performance is quantified using standard lexical metrics (BLEU-4~\cite{papineni2002bleu}, Rouge-L~\cite{lin2004rouge}) and radiology-specific metrics, including RadGraph~\cite{delbrouck2022improving} and CheXbert-based Macro-F1-14~\cite{chambon2024chexpert} (mapping ``uncertain'' labels to negative). 
For ReXGradient-160K, we additionally report RadCliQ~\cite{yu2023evaluating}, RaTEScore~\cite{zhao2024ratescore}, and BertScore~\cite{zhang2019bertscore}.

\begin{figure}[!t]
\centering
\includegraphics[width=0.90\linewidth]{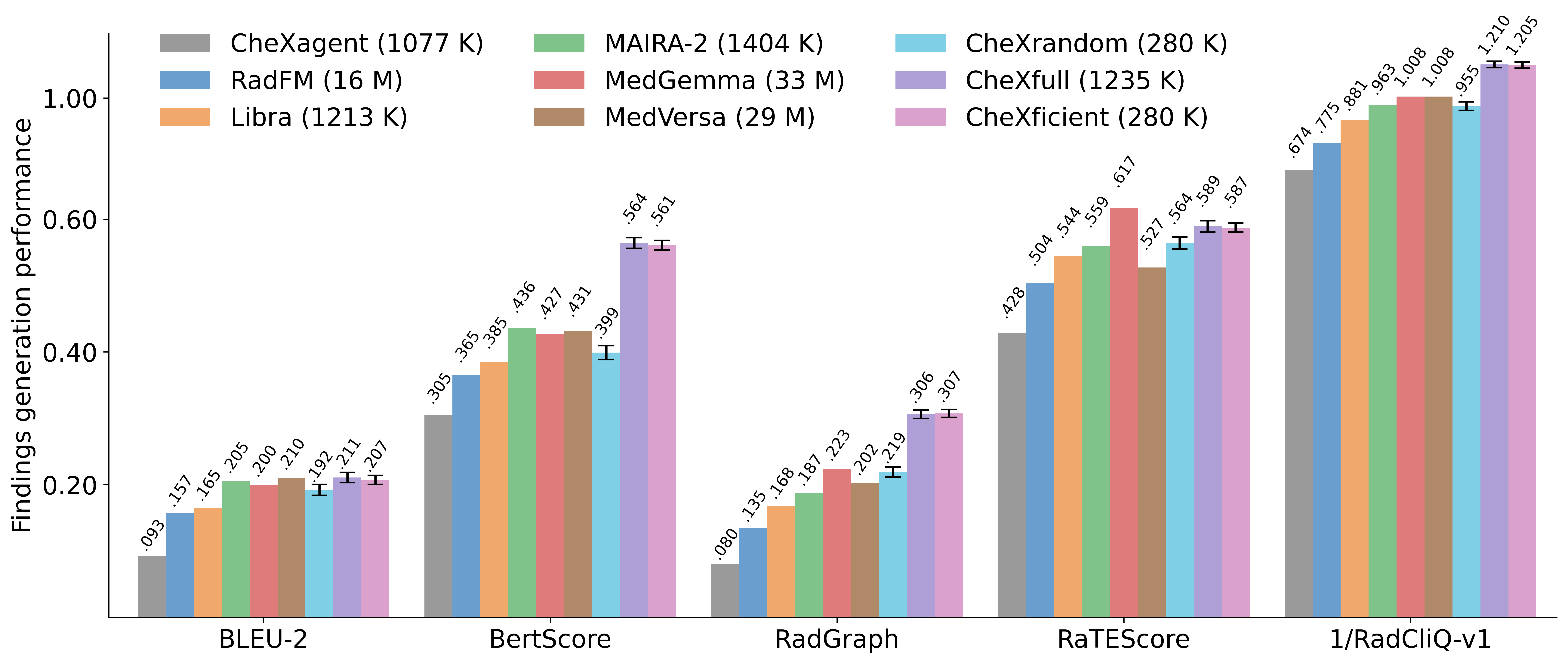}
\vspace{-5 pt}
\caption{
Downstream radiology report (\emph{Findings} section) generation performance (\%) on official test split of ReXGradient-160K ($n$ = 10,000). 
Text in parentheses denotes the pretraining data size of the model. 
}
\label{fig:report generation ReX160K}
% \vspace{-10pt}
\end{figure}

% \textbf{Result Analysis.}
CheXficient consistently outperforms CheXrandom across all lexical and clinical metrics on both MIMIC-CXR (Figure~\ref{fig:report generation MIMIC}) and ReXGradient-160K (Figure~\ref{fig:report generation ReX160K}).
Despite being pretrained on substantially fewer image–report pairs, CheXficient achieves performance comparable to CheXfull.
On MIMIC-CXR, CheXficient slightly under-performs the recent leading report generation framework LLaVA-Rad, while demonstrating improvements over other specialized vision–language baselines, including BiomedCLIP, CheXagent, and RAD-DINO~\cite{perez2025exploring}, all of which are pretrained on much larger-scale datasets (Extended Data Table~\ref{tab:comparative overview}).
On ReXGradient-160K, CheXficient further outperforms advanced medical foundation models, e.g., RadFM~\cite{wu2025towards}, Libra~\cite{zhang2025libra}, MAIRA-2~\cite{bannur2024maira}, and MedVersa~\cite{zhou2024medversa}.
% Overall, these results demonstrate that active curation during pretraining can substantially reduce data requirements while effectively preserving the quality of image representations for radiology report generation.

\subsection{Tackle long-tailed thoracic diseases}

% \caption{Impact of curated pretraining on per-disease generalization in VinDr-CXR.
% Top: disease label distributions in the VinDr-CXR training set (the y-axis uses a symmetric logarithmic scale to visualize label frequencies spanning multiple orders of magnitude).
% Bottom: corresponding zero-shot per-disease AUROC results on the VinDr-CXR test set.
% CheXficient shows different label distributions compared to CheXfull and CheXrandom, and achieves higher zero-shot AUROC for most disease categories.
% The disease label `edema' is not shown, as no samples with this label are present in the official test set.}

Chest radiograph interpretation is inherently a long-tailed problem~\cite{ma2025fully,wang2025cross}.
The distribution of thoracic findings is heavily skewed toward a small number of common abnormalities, as many clinically relevant conditions occur infrequently.
This severe class imbalance poses substantial challenges for model learning, as training objectives are often dominated by majority classes.
Consequently, AI models tend to exhibit biased representations that favor common findings, potentially at the expense of under-represented conditions.
Robustness and generalizability under long-tailed distributions therefore constitute critical criteria for evaluating medical foundation models.

\begin{figure}[!t]
\centering
\includegraphics[width=1.0\linewidth]{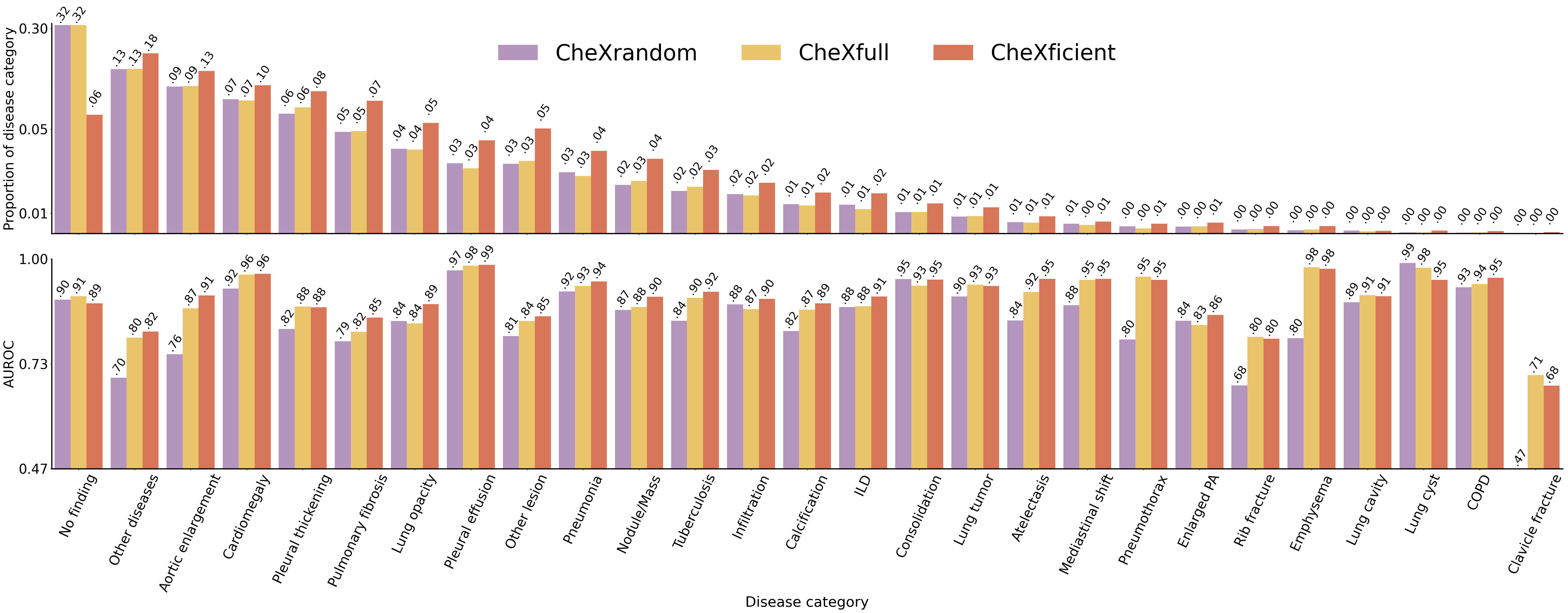}
\vspace{-15 pt}
\caption{Impact of curated pretraining on per-disease generalization in VinDr-CXR.
Top: disease label distributions from the VinDr-CXR training set (the y-axis uses a symmetric logarithmic scale to visualize disease label frequencies spanning multiple orders of magnitude). 
Bottom: corresponding zero-shot per-disease AUROC results on the VinDr-CXR test set. 
Curated pretraining (CheXficient) re-balances label distributions and yields improved zero-shot generalization for most disease categories compared to non-curated pretraining (CheXfull and CheXrandom).
Note that the disease label ‘edema’ is not shown in the figure as there are no samples with this label in the official test data.
}
\label{fig:VinDr-CXR}
% \vspace{-10pt}
\end{figure}

To investigate how active data curation during pretraining influences long-tailed generalization, we analyze the relationship between disease label distributions in the pretraining data and zero-shot generalization ability on test data.
We focus on VinDr-CXR~\cite{nguyen2022vindr}, as it provides comprehensive disease-level annotations covering a total of 28 findings and diagnoses, with long-tailed distributions.
Specifically, we first compute the disease label distributions in the VinDr-CXR training set (see Figure~\ref{fig:VinDr-CXR}(Top)).
Then, we compare the per-category zero-shot classification performance (AUROC) on the VinDr-CXR test data for different models (Figure~\ref{fig:VinDr-CXR}(Bottom))
% pretrained with full data (CheXfull), random subset data (CheXrandom), and curated data (CheXficient) in Figure~\ref{fig:VinDr-CXR}(Bottom).

We observe that \emph{No finding} samples dominate the non-curated training sets used by CheXfull and CheXrandom.
In contrast, CheXficient greatly reduces the proportion of \emph{No finding} samples through active curation, while almost preserving the generalization performance for this category on the test set.
Meanwhile, the curated pretraining set increases the relative prevalence of most disease labels, which is accompanied by improved zero-shot generalization across a broad range of pathologies (16 out of 26), such as \emph{Other diseases}, \emph{Aortic enlargement}, \emph{Pulmonary fibrosis}, \emph{Lung opacity}, \emph{Pleural effusion}, \emph{Other lesion}, and \emph{Pneumonia}.
However, this trend does not consistently hold for a small number of extremely-rare diseases, such as \emph{Lung cyst} and \emph{Clavicle fracture}, which each account for less than 0.1\% of the training data.
We hypothesize that, despite being relatively up-weighted by active curation, the absolute number of samples for these conditions remains insufficient to support robust representation learning.
This observation suggests a potential limitation of data curation strategies alone in addressing long-tailed distributions and highlights that complementary approaches, such as targeted data acquisition or task-specific augmentation, may be necessary for extremely-rare disease modeling.

\begin{figure}[!t]
\centering
\includegraphics[width=1.0\linewidth]{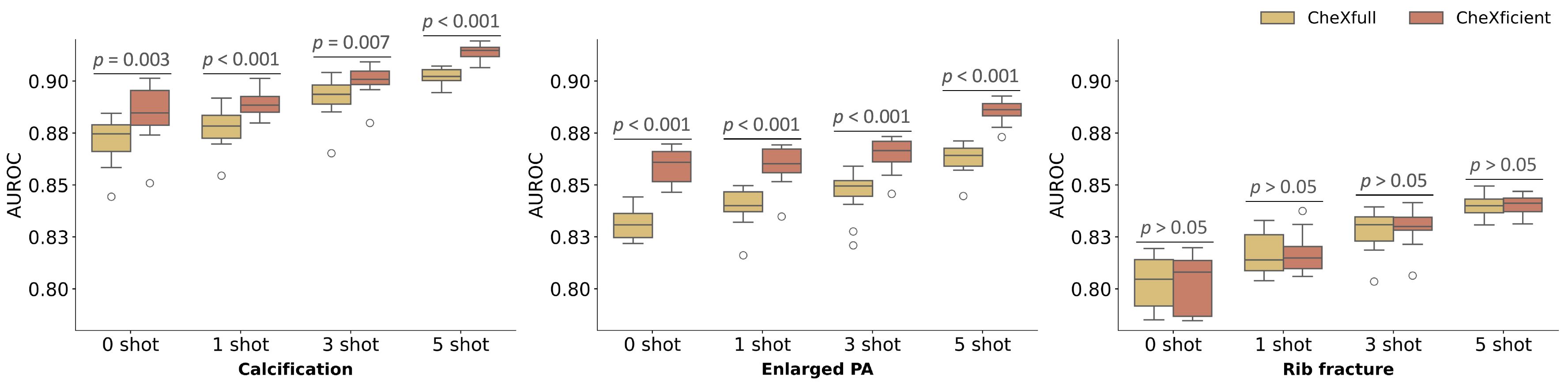}
\vspace{-15 pt}
\caption{Adaptability of CheXficient in detecting long-tailed conditions (Calcification, Enlarged PA (pulmonary artery), and Rib fracture) under a two-way (no finding versus each long-tailed condition) $k$-shot learning setting.
The results demonstrate the ability of CheXficient to adapt to data-scarce conditions using only a few labeled training samples.
Box plots show AUROC distributions over 20 experimental replicates, with the median (central line), interquartile range (box indicating the 25th and 75th percentiles), whiskers extending to 1.5 times the interquartile range, and outliers shown as individual points.
$p$-values from two-sided $t$-tests are reported.
}
\label{fig:Longtailed_VinDr-CXR}
% \vspace{-10pt}
\end{figure}

We further examine the adaptability of CheXficient to long-tailed conditions under limited supervision by finetuning the pretrained models with a linear-probing configuration, using only 1–5 labeled samples ($k$-shot learning)~\cite{ma2025fully}.
Figure~\ref{fig:Longtailed_VinDr-CXR} presents box plots of AUROC distributions for three representative VinDr-CXR conditions.
% Both models exhibit outliers with AUROC scores below 50\%, reflecting the intrinsic difficulty of detecting long-tailed conditions with extremely limited supervision.
CheXficient outperforms CheXfull, achieving higher median and maximum AUROC scores for \emph{Calcification} and \emph{Enlarged PA}, and comparable performance for \emph{Rib fracture}.
These results demonstrate that CheXficient can leverage scarce labeled data to detect long-tailed conditions, underscoring its practical value in data-scarce environments.

% Aortic enlargement, (2) Atelectasis, (3) Cardiomegaly, (4) Calcification, (5) Clavicle fracture, (6) Consolidation, (7) Edema, (8) Emphysema, (9) Enlarged PA, (10) Interstitial lung disease (ILD), (11) Infiltration, (12) Lung cavity, (13) Lung cyst, (14) Lung opacity, (15) Mediastinal shift, (16) Nodule/Mass, (17) Pulmonary fibrosis, (18) Pneumothorax, (19) Pleural thickening, (20) Pleural effusion, (21) Rib fracture, (22) Other lesion, (23) Lung tumor, (24) Pneumonia, (25) Tuberculosis, (26) Other diseases, (27) Chronic obstructive pulmonary disease (COPD), and (28) No finding.

\section{Discussion}
\label{discussion}

% While recent work has explored synthetic data generation~\cite{bluethgen2025vision,sun2025data} to complement training samples, 
% such approaches introduce additional computational overhead and raise concerns about fidelity and clinical validity. 

% To address data scarcity, prior studies~\cite{bluethgen2025vision,sun2025data} have explored generative AI techniques~\cite{rombach2022high} to synthesize visually plausible medical training samples. However, such approaches are inherently constrained by the fidelity and clinical reliability of generated data, while simultaneously incurring additional computational costs due to the inclusion of large volumes of synthetic samples. 

In this study, we propose a data- and compute-efficient pretraining approach for medical foundation models that leverages active, principled, and intelligent data curation.
Rather than indiscriminately enlarging training datasets~\cite{bluethgen2025vision,sun2025data} and incurring substantial computational costs, our approach optimizes the utilization of data and compute budgets during pretraining, offering a cost-effective alternative for building high-performing medical foundation models.
By curating 280K image–report pairs (22.7\%) from a pool of 1.235 million paired CXR images and radiology reports, we pretrained a new CXR foundation model, CheXficient.
Extensive experiments demonstrate that CheXficient achieves comparable or superior performance to a full-data pretrained counterpart across diverse evaluation benchmarks spanning 5 evaluation protocols.
These evaluations cover both non-adapted tasks, including zero-shot findings classification and zero-shot cross-modal retrieval, as well as adapted downstream tasks such as multi-label disease prediction, semantic segmentation of anatomical structures and abnormalities, and radiology report generation.
Together, these results provide strong evidence that curated pretraining on large-scale medical datasets can substantially improve data and compute efficiency as well as generalizability, without sacrificing overall performance.
We also analyze feature distributions of the curated training samples (Figure~\ref{fig:feat_distribution}).
The curated data exhibits distinctive local sample-density structures, reflecting its ability to preserve data diversity while emphasizing under-represented and informative samples.
This structural property is closely associated with strong model performance (Figure~\ref{fig:zero-shot classification} and Figure~\ref{fig:zero-shot retrieval}) and improved long-tailed generalization (Figure~\ref{fig:VinDr-CXR} and Figure~\ref{fig:Longtailed_VinDr-CXR}) in the resulting CXR foundation model.

In the pretraining phase, CheXficient integrates an online data curator that employs a set of evolving prototypes (i.e., prototypical centroids) to approximate the global data manifold.
The curator selectively prioritizes informative CXR image–report pairs for model optimization: samples that lie far from the learned prototypes are emphasized, while those close to the prototypes are down-weighted and under-sampled.
Those unselected samples are excluded from the pretraining process.
In comparison with CheXfull, which shares an identical network architecture but is pretrained on the entire training set of 1.235 million data pairs, CheXficient achieves substantially higher pretraining data efficiency (Figure~\ref{fig:pretraining data efficiency}).
In addition, CheXficient requires significantly less pretraining computational cost than CheXfull (Figure~\ref{fig:pretraining time efficiency}).

Despite being pretrained on only 280K image–report pairs, CheXficient demonstrates strong and consistent performance across a wide range of downstream applications, including both unimodal evaluations (image classification and semantic segmentation) and multimodal tasks such as radiology report generation.
Across eight datasets, CheXficient achieves performance comparable to or exceeding that of CheXfull (Figures~\ref{fig:classification label efficiency}, \ref{fig:segmentation annotation efficiency}, \ref{fig:report generation MIMIC}, and \ref{fig:report generation ReX160K}).
Another appealing capability of CheXficient is its ability to reduce the amount of expert-annotated training data required for downstream tasks.
Figure~\ref{fig:classification label efficiency} demonstrates the high labeling efficiency of CheXficient in terms of image classification.
On the MIMIC-CXR dataset, it achieves the classification performance of CheXrandom (pretrained on a random subset with an identical size as the curated subset), using less than 10\% of the labeled finetuning data. 
A similar trend is observed for downstream semantic segmentation tasks (Figure~\ref{fig:segmentation annotation efficiency}).
Furthermore, CheXficient shows highly competitive results of radiology report generation compared with other large-scale pretrained models (Figure~\ref{fig:report generation MIMIC} and Figure~\ref{fig:report generation ReX160K}).

We further conducted analyses on patient demographic characteristics, including age and race.
We observed that the curated subset contains a relatively higher proportion of training samples from two age groups: pediatric patients (0–5 years), whose CXRs exhibit greater anatomical variability, and older adults (over 65 years), whose CXRs often present a broader spectrum of age-related pathologies due to their higher prevalence in this population (Appendix~\ref{age}).
Regarding patient race, the data curation process does not exhibit noticeable bias, and the racial distribution remains comparable between the curated subset and the full training set (see Appendix~\ref{race}).
In addition, the curation process exhibits a mild preference for frontal CXR scans over lateral CXR scans (Appendix~\ref{frontal_lateral}), as frontal CXRs may provide more informative visual cues for many radiological findings.
It also demonstrates a pronounced preference for samples with longer CXR reports (Appendix~\ref{text_token}), which are prone to contain richer and more detailed clinical descriptions.
Combined with the observations from Figure~\ref{fig:VinDr-CXR}, these results inspire an important and broad implication: data acquisition or collection practices for developing medical foundation models may benefit from prioritizing under-represented data, long-tailed populations, and informative samples with rich visual and textual clinical content.

Despite promising results, we acknowledge several limitations in our methodology and evaluations.
First, due to the limited scope of our study, we have not studied alternative encoder architecture adaptations, e.g., ViT-Large architecture adopted by other CXR foundation models (Extended Data Table~\ref{tab:comparative overview}). 
We expect that incorporating more advanced encoder architectures into our framework could yield additional performance gains due to enhanced representational capacity, and we leave this exploration for future work.
Second, our multimodal evaluation is currently limited to textual report generation.
Other important radiological multimodal tasks, such as visual question answering~\cite{hu2024interpretable,liu2025gemex} and visual grounding~\cite{huy2025seeing}, remain unexplored and represent promising directions for future research.
Third, although our validation covers 47 thoracic findings and diseases across major categories (e.g., cardiomegaly, pneumonia, and nodule), this constitutes only a fraction of known thoracic conditions (for example, over 170 radiographic findings reported in~\cite{bustos2020padchest}), with limited coverage of rare diseases and complex systemic diseases. 
Future studies should aim for broader and more comprehensive evaluation across a wider spectrum of thoracic conditions.

In summary, we present a data- and compute-efficient CXR foundation model based on active, principled, and intelligent data curation.
Under non-adapted evaluation settings, CheXficient demonstrates strong zero-shot generalization performance while achieving high pretraining data and computation efficiency.
When adapted to a range of downstream tasks, CheXficient consistently exhibits high labeling and fine-tuning efficiency without compromising performance.
Overall, our work highlights the potential for substantial cost savings through strategic dataset acquisition that prioritizes informative and long-tailed data, and provides valuable insights for medical domains where the development of foundation models has been constrained by limited access to large-scale annotated data and extensive computational resources.

\section{Methods}
\label{method}

\subsection{Datasets for pretraining}

We construct a large-scale pretraining corpus comprising over 1.235 million paired CXR–report samples, which are collected and aggregated from 13 publicly available datasets covering diverse patient populations worldwide, including MIMIC-CXR~\cite{johnson2019mimic}, CheXpert Plus~\cite{chambon2024chexpert}, ReXGradient-160K~\cite{zhang2025rexgradient}, PadChest~\cite{bustos2020padchest}, BIMCV COVID-19~\cite{vaya2020bimcv}, CANDID-PTX~\cite{feng2021curation}, CASIA-CXR~\cite{metmer2024open}, Open-I~\cite{demner2015preparing}, ChestX-ray14~\cite{wang2017chestx}, BRAX~\cite{reis2022brax}, VinDr-CXR~\cite{nguyen2022vindr}, VinDr-PCXR~\cite{pham2023pedicxr}, and ChestDR~\cite{wang2023real}.
The detailed composition and distribution of the full dataset are summarized in Table~\ref{tab:pretraining_data_statistics}.
This diverse aggregation enables broad coverage of imaging protocols, disease spectra, and textual reporting styles, which is critical for robust and generalizable representation learning.
To ensure transparency and reproducibility, we strictly follow the official train/validation/test splits provided by each dataset when available, and incorporate only the train split into the pretraining corpus. 

Among the 13 contributing datasets, 8 provide raw radiology reports, including MIMIC-CXR~\cite{johnson2019mimic}, CheXpert Plus~\cite{chambon2024chexpert}, ReXGradient-160K~\cite{zhang2025rexgradient}, PadChest~\cite{bustos2020padchest}, BIMCV COVID-19~\cite{vaya2020bimcv}, CANDID-PTX~\cite{feng2021curation}, CASIA-CXR~\cite{metmer2024open}, and Open-I~\cite{demner2015preparing}. Notably, several of these datasets (PadChest, BIMCV COVID-19, and CASIA-CXR) contain non-English reports, which are automatically translated into English to ensure linguistic consistency across the corpus. 
For all datasets with reports, regex-based filtering is applied to handle common formatting inconsistencies such as repeated whitespace and malformed line breaks.
Samples in which both the \emph{Findings} and \emph{Impression} sections are blank are excluded from the corpus.

The remaining 5 datasets do not include free-text reports but instead provide structured diagnostic labels (e.g., pleural effusion, cardiomegaly, atelectasis). 
They are ChestX-ray14~\cite{wang2017chestx}, BRAX~\cite{reis2022brax}, VinDr-CXR~\cite{nguyen2022vindr}, VinDr-PCXR~\cite{pham2023pedicxr}, and ChestDR~\cite{wang2023real}. 
For these datasets, we generate pseudo-reports using a template-based report synthesis strategy introduced in prior work~\cite{zambrano2025clinically}. 
The resulting synthetic reports exhibit a token-length distribution comparable to that of real-world reports (Table~\ref{tab:pretraining_data_statistics}) and account for only a small portion (12.6\%) of the overall pretraining corpus.

\begin{table*}[h]
\centering
\setlength{\extrarowheight}{0pt}
\addtolength{\extrarowheight}{\aboverulesep}
\addtolength{\extrarowheight}{\belowrulesep}
\setlength{\aboverulesep}{3pt}
\setlength{\belowrulesep}{0pt}
\caption{
An extensive corpus of 1.235 million CXR image-report pairs employed for model pretraining in our experiments. 
We exclude images that do not belong to standard views (i.e., frontal or lateral).
Real-world \emph{Findings} are extracted from the free-text radiology reports.
Synthetic \emph{Findings} are generated using template sentences derived from clinical diagnostic labels. 
}
\vspace{-6pt}
\label{tab:pretraining_data_statistics}

\setlength{\tabcolsep}{2.0 mm}
\resizebox{1.0\linewidth}{!}{

\begin{tabular}{lccccccc}
\hline
\multirow{2}{*}{Source dataset} & \multirow{2}{*}{Country} & \multirow{2}{*}{Patients} & \multirow{2}{*}{Labels} & \multicolumn{3}{c}{Text tokens} & \multirow{2}{*}{Image-text pairs} \\
\cmidrule(l){5-7}
                                &                          &                            &                        & 5\%-ile & Average & 95\%-ile &                         \\
\hline
\multicolumn{8}{l}{\textit{Real-world Findings (87.4\%)}} \\
\hline
MIMIC-CXR        & U.S.        & 65,379 & –  & 42 & 94  & 176 & 354,619 \\
ReXGradient-160K & U.S.        & 95,716 & –  & 35 & 76  & 152 & 238,965 \\
CheXpert-Plus    & U.S.        & 64,725 & –  & 38 & 98  & 209 & 222,116 \\
PadChest         & Spain       & 67,625 & –  & 6  & 30  & 76  & 160,672 \\
BIMCV-COVID19    & Spain       & 25,082 & –  & 8  & 46  & 102 & 65,421  \\
CANDID-PTX       & New Zealand & 13,744 & –  & 8  & 42  & 97  & 19,609  \\
CASIA-CXR        & France      & 11,111 & –  & 57 & 96  & 145 & 11,111  \\
Open-I           & U.S.        & 3,996  & –  & 29 & 72  & 139 & 7,424   \\
\hline
\multicolumn{8}{l}{\textit{Synthetic Findings (12.6\%)}} \\
\hline
NIH ChestX-ray14 & U.S.     & 28,008 & 15 & 17 & 72  & 159 & 86,524 \\
BRAX             & Brazil   & 18,442 & 14 & 11 & 29  & 65  & 40,967 \\
VinDr-CXR        & Vietnam  & 15,000 & 28 & 17 & 107 & 325 & 15,000 \\
VinDr-PCXR       & Vietnam  & 7,728  & 15 & 17 & 72  & 189 & 7,728  \\
ChestDR          & China    & 4,848  & 19 & 20 & 122 & 239 & 4,848  \\
\hline
\multicolumn{7}{r}{Total:} & 1,235,004 \\
\hline
\end{tabular}
}
\end{table*}

\subsection{Datasets for non-adapted zero-shot tasks}

% TBX11K~\cite{liu2020rethinking} contains 11,200 CXR images with image-level tuberculosis (TB) annotations. We use it for binary TB classification.
% SIIM-PTX~\cite{SIIM-ACR-Pneumothorax-Segmentation-2019} comprises over 12k frontal-view CXR images with pneumothorax (PTX) masks, used for both PTX classification and segmentation tasks.
% Pneumonia2017~\cite{kermany2018labeled} includes 5,856 pediatric CXR images, which are used for binary pneumonia classification in our experiment.
% VinDr-RibCXR~\cite{nguyen2021vindr} is a rib segmentation dataset containing 245 frontal-view CXR images (196 for training and 49 for testing), used for binary rib segmentation.

As shown in Table~\ref{tab:evaluation_data_statistics}, non-adapted zero-shot findings classification is evaluated on the official test sets of 8 datasets: MIMIC-CXR~\cite{johnson2019mimic}, CheXpert~\cite{chambon2024chexpert}, ChestX-ray14~\cite{wang2017chestx}, VinDr-CXR~\cite{nguyen2022vindr}, VinDr-PCXR~\cite{pham2023pedicxr}, TBX11K~\cite{liu2020rethinking}, SIIM-PTX~\cite{SIIM-ACR-Pneumothorax-Segmentation-2019}, and Pneumonia2017~\cite{kermany2018labeled}.
Since the pretraining pool assembles multiple public datasets, some of which (MIMIC-CXR, CheXpert, ChestX-ray14, VinDr-CXR, VinDr-PCXR) are also used for evaluation, we strictly follow the official test splits for these datasets, ensuring patient-level separation between pretraining and evaluation.
This precaution addresses potential data leakage, which is the first consideration in our experiments.
Notably, SIIM-PTX~\cite{SIIM-ACR-Pneumothorax-Segmentation-2019}, Pneumonia2017~\cite{kermany2018labeled}, and TBX11K~\cite{liu2020rethinking} are unseen out-of-distribution datasets, with their test sets completely disjoint from the pretraining sources, and are therefore treated as external evaluation.
Following the zero-shot classification protocol in CheXzero~\cite{tiu2022expert}, for each disease label (e.g., mass, edema, atelectasis), we leverage the prompts “$\langle$ disease $\rangle$” and “no $\langle$ disease $\rangle$” as positive and negative prompts for the text encoder, obtaining softmax-based probabilistic predictions.

For the non-adapted zero-shot cross-modal retrieval, evaluation is performed on two datasets: MIMIC-CXR and CheXpert.
On MIMIC-CXR, the standard test split is utilized.
Since CXR reports are unavailable for the CheXpert test split, we instead use CheXpert5x200~\cite{huang2021gloria,you2023cxr} for the retrieval evaluation.
To prevent data leakage, all the 1000 paired CXR samples in CheXpert5x200 are excluded from the pretraining pool in advance, ensuring a fair and disjoint evaluation set.

\begin{table*}[h]
\centering
\setlength{\extrarowheight}{0pt}
\addtolength{\extrarowheight}{\aboverulesep}
\addtolength{\extrarowheight}{\belowrulesep}
\setlength{\aboverulesep}{3pt}
\setlength{\belowrulesep}{0pt}
\caption{
Summary of the evaluation datasets used in this study. We evaluate on a total of 11 datasets (6 seen datasets for internal evaluation and 5 unseen datasets for external evaluation), comprising diverse evaluation benchmarks across 5 task types. 
We follow the official train/test splits provided by each dataset, except for CheXpert, where 1,000 paired CXR samples were excluded from the training set for retrieval evaluation.
% *CXR-LT expands on MIMIC-II with 12 new classes, among which three rare diseases are selected for our few-shot learning. To ensure no samples are seen by CheXficient during its pretraining, only images from the hold-out validation and test sets of MIMIC-II were utilized.
}
\vspace{-6pt}
\label{tab:evaluation_data_statistics}

\setlength{\tabcolsep}{1.5 mm}
\resizebox{1.0\linewidth}{!}{

\begin{tabular}{lccccc} 
\toprule
Dataset                                                                                              & Learning setup & Task              & Clinical conditions       & \# Train & \# Test  \\ 
\midrule
\multicolumn{6}{l}{\textit{Internal evaluation}}                                                                                                                                            \\ 
\hline
\rowcolor[rgb]{0.930,0.930,0.930} {\cellcolor[rgb]{0.930,0.930,0.930}}                               & Non-adapted    & Classification    & 14 thoracic findings      & 0        & 3,082    \\
\rowcolor[rgb]{0.930,0.930,0.930} {\cellcolor[rgb]{0.930,0.930,0.930}}                               & Non-adapted    & Retrieval         & –                         & 0        & 3,082    \\
\rowcolor[rgb]{0.930,0.930,0.930} {\cellcolor[rgb]{0.930,0.930,0.930}}                               & Adapted        & Classification    & 14 thoracic findings      & 227,459 (1–100\%) & 3,082    \\
\rowcolor[rgb]{0.930,0.930,0.930} \multirow{-4}{*}{{\cellcolor[rgb]{0.930,0.930,0.930}}MIMIC-CXR}    & Adapted        & Report generation & –                         & 146,909  & 2,461    \\
\multirow{3}{*}{CheXpert}                                                                            & Non-adapted    & Classification    & 14 thoracic findings      & 0        & 500      \\
                                                                                                     & Non-adapted    & Retrieval         & –                         & 0        & 1,000    \\
                                                                                                     & Adapted        & Classification    & 14 thoracic findings      & 223,414 (1–100\%)  & 500      \\
\rowcolor[rgb]{0.930,0.930,0.930} {\cellcolor[rgb]{0.930,0.930,0.930}}                               & Non-adapted    & Classification    & 15 thoracic findings      & 0        & 25,596   \\
\rowcolor[rgb]{0.930,0.930,0.930} \multirow{-2}{*}{{\cellcolor[rgb]{0.930,0.930,0.930}}NIH ChestX-ray14} & Adapted        & Classification    & 15 thoracic findings      & 86,524 (1–100\%)   & 25,596   \\
VinDr-CXR                                                                                            & Non-adapted    & Classification    & 28 thoracic findings      & 0, $k$-shot        & 3,000    \\
VinDr-PCXR                                                                                           & Non-adapted    & Classification    & 15 thoracic findings      & 0        & 1,397    \\
ReXGradient-160K                                                                                     & Adapted        & Report generation & –                         & 140,000  & 10,000  \\
% CXR-LT*                                                                                              & Non-adapted    & Classification    & Emphysema, Tortuous aorta & 0        & ?        \\ 
\hline
\multicolumn{6}{l}{\textit{External evaluation}}                                                                                                                                            \\ 
\hline
\rowcolor[rgb]{0.930,0.930,0.930} {\cellcolor[rgb]{0.930,0.930,0.930}}                               & Non-adapted    & Classification    & Pneumothorax              & 0        & 1,372    \\
\rowcolor[rgb]{0.930,0.930,0.930} \multirow{-2}{*}{{\cellcolor[rgb]{0.930,0.930,0.930}}SIIM-PTX}     & Adapted        & Segmentation      & Pneumothorax              & 10,675 (1–100\%)   & 1,372    \\
Pneumonia2017                                                                                        & Non-adapted    & Classification    & Pediatric pneumonia       & 0        & 624      \\
TBX11K                                                                                               & Non-adapted    & Classification    & Tuberculosis              & 0        & 1,800    \\
JSRT                                                                                                 & Adapted        & Segmentation      & Lungs                     & 172 (1–100\%)      & 50       \\
VinDr-RibCXR                                                                                         & Adapted        & Segmentation      & Ribs                      & 196 (1–100\%)      & 49       \\
\bottomrule
\end{tabular}

}
\end{table*}

\subsection{Datasets for adapted downstream tasks}

For adapted downstream classification tasks, evaluations are performed on the official test sets of MIMIC-CXR, CheXpert, and ChestX-ray14 (Table~\ref{tab:evaluation_data_statistics}), as in the zero-shot setting. 
The key difference is that varying percentages (1–100\%) of data from the training split are used from each dataset to fine-tune the linear classification head.

For adapted downstream image segmentation tasks, the SIIM-PTX~\cite{SIIM-ACR-Pneumothorax-Segmentation-2019}, JSRT~\cite{shiraishi2000development}, and VinDr-RibCXR~\cite{nguyen2021vindr} datasets are employed for fine-tuning and evaluation, 
with the official data splits summarized in Table~\ref{tab:evaluation_data_statistics}.
% SIIM-PTX comprises about 12,000 frontal-view CXR images with corresponding pneumothorax (PTX) masks (10,675 training / 1,372 testing), and is used for binary PTX segmentation.
% VinDr-RibCXR is composed of 245 frontal-view CXRs with pixel-wise rib masks (196 training / 49 testing), and is utilised for binary segmentation of ribs.

For downstream-adapted radiology report generation tasks, fine-tuning and evaluation are performed on the official splits of MIMIC-CXR and ReXGradient-160K.
Following the protocols of LLaVA-Rad~\cite{zambrano2025clinically} and RAD-DINO~\cite{perez2025exploring}, non-frontal images and samples without a \emph{Findings} section are excluded, forming 146,909 training, 7,250 validation, and 2,461 testing image–text pairs for MIMIC-CXR.
A similar pre-processing pipeline is applied to ReXGradient-160K, yielding 140,000, 10,000, and 10,000 image–text pairs for training, validation, and testing, respectively.

\subsection{Active data curation in pretraining}
\label{sec:algorithm}

Our CheXficient adopts a contrastive language–image pretraining framework based on CLIP~\cite{radford2021learning} to actively curate informative and diverse training samples from a large-scale corpus containing over 1.235M paired CXR image–report samples. 
Unlike conventional pretraining pipelines that treat all samples equally, CheXficient performs online data curation during training, dynamically identifying samples that are most beneficial for the representation learning. This strategy enables efficient utilization of large-scale medical datasets, which are often redundant, imbalanced, and noisy, while substantially reducing computational cost.
Formally, we consider a large image–text dataset $\mathcal{D} = \{ ({x}_i^{\text{img}}, {x}_i^{\text{txt}}) \}_{i=1}^{|\mathcal{D}|}$, where each data pair consists of a CXR image ${x}_i^{\text{img}}$ and its corresponding textual radiology report ${x}_i^{\text{txt}}$.
Our objective is to construct a compact yet representative subset $\mathcal{D}_c \subset \mathcal{D}$ during pretraining, such that a vision–language model trained on $\mathcal{D}_c$ achieves performance comparable to training on the full dataset $\mathcal{D}$, but with significantly fewer samples and lower training cost.

The overall curation and pretraining procedure consists of the following steps:
\begin{enumerate}[
    label=(\arabic*),
    leftmargin=3em,
    labelsep=0.5em,
    align=left
]
    \item \textbf{Multimodal embedding.}
    During pretraining, we randomly sample a large super-batch of paired CXR data $\mathcal{S} = \{ ({x}_i^{\text{img}}, {x}_i^{\text{txt}}) \}_{i=1}^{|\mathcal{S}|}$ from the full pretraining set $\mathcal{D}$. 
    Each radiology report $x_i^{\text{txt}}$ is tokenized and encoded by the text encoder $\theta_{\text{txt}}$ to obtain a text embedding $\mathbf{z}_i^{\text{txt}}$.
    Correspondingly, each CXR image $x_i^{\text{img}}$ is resized and processed by the image encoder $\theta_{\text{img}}$ to extract an image embedding $\mathbf{z}_i^{\text{img}}$ of the same dimensionality.
    Both embeddings are $\ell_2$-normalized to ensure scale consistency and then concatenated to form a unified multimodal representation $\mathbf{z}_i = {\mathsf{concat}}(\mathbf{z}_i^{\text{img}}, \mathbf{z}_i^{\text{txt}})$, capturing complementary semantic information from both visual and textual modalities (see ablations in Appendix~\ref{ablation_embeddings}), 
    which is used for subsequent data selection. 

    \item \textbf{Prototype-driven selection.} 
    The model maintains a set of evolving prototypes $\{ \mathbf{p}_k \}_{k=1}^{K}$ that characterize representative multimodal patterns in the data manifold (Figure~\ref{fig:workflow}(b)). 
    For each concatenated embedding in $\mathcal{S}$, we compute its distance to the nearest prototype and rank samples accordingly.  
    Samples that lie far from their assigned prototypes are more likely to correspond to rare pathologies, atypical imaging appearances, or under-represented report semantics, and are therefore highly informative.  
    Accordingly, we retain a small fraction (e.g., the top 10\%) of these distant samples to emphasize hard and informative examples.  
    For the remaining majority of samples (e.g., 90\%), which tend to be redundant and clustered around common patterns, we under-sample a fixed number of data (e.g., $N=10$) within each prototype cluster to preserve diversity. 
    This under-sampling step is implemented using farthest point sampling~\cite{eldar1997farthest}, enhancing coverage of the feature space.
    
    \item \textbf{Contrastive learning.}  
    The selected samples constitute a mini-batch $\mathcal{M} \subset \mathcal{S}$, which is added to the curated subset $\mathcal{D}_c$.  
    Given the mini-batch $\mathcal{M}$, we optimize the model using the InfoNCE contrastive loss~\cite{oord2018representation}, jointly updating the image and text encoders.  
    This objective encourages matched CXR image–report pairs to have aligned representations, while pushing apart mismatched pairs, thereby improving cross-modal alignment and semantic consistency.

    \item \textbf{Prototype update.}  
    The prototypes are updated concurrently with model training to reflect the evolving data distribution. 
    Specifically, new prototype representations are computed as weighted averages of the concatenated embeddings from the selected samples, where weights are determined by each sample’s embedding proximity to the corresponding prototype.  
    We achieve this via an optimal transport based clustering algorithm~\cite{cuturi2013sinkhorn} that can enforce unique assignment and equipartition constraints across prototypes.
    % To ensure balanced assignments and prevent prototype collapse, we employ an optimal transport–based clustering algorithm~\cite{cuturi2013sinkhorn}, which enforces unique assignment and equipartition constraints across prototypes. 
    To further stabilize prototype estimation and incorporate historical information, we apply an exponential moving average (EMA)~\cite{hunter1986exponentially} update scheme.

\end{enumerate}

These above steps are iteratively executed throughout the curated pretraining process, enabling the model to progressively identify and emphasize informative, diverse, and under-represented samples. 
The data-curated pretraining procedure is summarized in Extended Data Algorithm~\ref{alg:training_algorithm}. 
After pretraining, the resulting model can be directly applied to zero-shot evaluation in a non-adapted scheme or further fine-tuned with task-specific heads or decoders for a wide range of downstream tasks including classification, segmentation, and textual report generation.

\subsection{Implementation details}

CheXficient, CheXfull, and CheXrandom share the same model architecture, training configuration, and hardware setup, and all models are trained for a total of 20 epochs for a fair comparison.
The architecture follows a specific instantiation of the CLIP framework, consisting of an image encoder and a text encoder.
We adopt DINOv2~\cite{oquab2023dinov2} as the image encoder, using the $\mathsf{[CLS]}$ token representation, and BioClinicalBERT~\cite{alsentzer2019publicly} as the text encoder.
Both encoders are followed by a linear projection layer that maps image and text features into a shared embedding space of 512 dimensions.
DINOv2 is implemented as a ViT-Base~\cite{dosovitskiy2020image} model with 12 transformer blocks (approximately 86 million parameters), while BioClinicalBERT consists of 12 transformer layers with about 110 million parameters.

During pretraining, CXR images are resized to $378 \times 378$ before being fed into the image encoder.
No data augmentation is applied to either image or text inputs.
For the textual modality, the \emph{Findings} and \emph{Impression} sections of each radiology report are combined (when both are available), and used as input to the text encoder, with a maximum context length of 256 tokens.
Model training is driven by an InfoNCE contrastive loss with a learnable temperature parameter, which is initialized to $0.01$.
The InfoNCE loss encourages alignment between paired CXR images and reports while separating mismatched pairs.
The number of prototypes $K$ = 6, with a sensitivity analysis provided in Appendix~\ref{ablation_prototype_number}. 
Before starting the curation process, we warm up the prototypes by initializing them using k-means centroids~\cite{hartigan1979algorithm} computed from $6,400$ training samples.
In the EMA update of prototypes, a smoothing factor of $0.1$ is applied.
In experiments, we observe that a single training epoch suffices to identify and curate representative samples from the full pretraining corpus,
after which the curated subset is reused for the remaining epochs for improving efficiency. 
% as subsequent curation passes yield highly overlapping subsets with negligible performance gain, indicating that the selected samples remain representative as the model evolves. 
Optimization is conducted using the AdamW~\cite{loshchilov2017fixing} optimizer with an initial learning rate of $5 \times 10^{-5}$, a weight decay of $1 \times 10^{-4}$, and a cosine annealing learning rate schedule.
Since the most distant samples from each prototype may contain noise or outliers, e.g., corrupted images, severe artifacts, report translation errors, image–report mismatches, or template-generated reports with weak visual grounding, we first remove these extreme outliers (i.e., 5\%) within each super-batch before performing the main curation step.
Each super-batch contains $640$ paired samples (8 GPUs $\times$ 80 pairs per GPU). 
Notably, we find that CheXficient can also be trained on a single GPU by leveraging an embedding accumulation strategy, 
in which samples are repeatedly drawn from the pretraining set, their feature embeddings are extracted sequentially, and then accumulated to construct a large super-batch.
This strategy can effectively enlarge the super-batch size without increasing the GPU count, enabling memory-efficient training while preserving the benefits of large-batch curation. 

% \chong{mention negligible overhead of prototype computing, ordering and searching, and prototype updating.}
In fact, CheXficient introduces negligible additional pretraining overhead in comparison to each base model. 
The potentially increased overhead involves two major steps: 
1) sample-prototype distance calculation: candidate training samples are drawn from in-batch data, which typically consists of only a few hundred samples, enabling efficient distance calculations to prototypes in the embedding space;
2) prototype update: new prototypes are obtained as weighted averages of embeddings from the selected samples and updated using an EMA scheme. This operation incurs minimal computational cost, as confirmed by our empirical observations.

% $\{\mathbf{p}_m\}^{t+1}_{\rm ema} := \tau \{\mathbf{p}_m\}^{t}_{\rm ema} + (1 - \tau) \{\mathbf{p}_m\}^{t}$, where $\tau = 0.9$

\subsection{Evaluation metrics}

Classification performance is evaluated using well-established metrics widely adopted in prior studies~\cite{chen2024chexagent,tiu2022expert,perez2025exploring}, including the area under the receiver operating characteristic curve (AUROC) and the area under the precision–recall curve (AUPRC). 
AUROC characterizes the trade-off between the true positive rate (recall) and the false positive rate across varying decision thresholds, while AUPRC captures the trade-off between precision and recall.
For multi-label classification, AUROC and AUPRC are computed independently for each disease category and then averaged to obtain the macro AUROC and macro AUPRC.
Evaluation of retrieval tasks leverages the metric of Recall@1, which measures whether the exact paired radiology report is retrieved within the top-1 result for a given CXR image, and vice versa.
Segmentation performance is evaluated using the Dice score, which quantifies pixel-wise overlap between the predicted segmentation maps and the corresponding ground-truth annotations.
Performance of radiology report generation is evaluated using standard lexical metrics (BLEU-2, BLEU-4, and Rouge-L) which measure word-level overlap between the generated and reference reports, general semantic similarity metrics (BERTScore) that measure contextual embedding-level semantic similarity between generated and reference reports, and radiology-specific metrics (RadGraph, Macro-F1-14, RaTEScore, and RadCliQ) which assess the clinical correctness and factual consistency of the generated reports.

\subsection{Statistical analysis}

To guarantee reliability and robustness of this study, we trained the model with five different random seeds and calculated the mean and standard deviation of the performance metrics, from which the standard error is computed as the standard deviation divided by $\sqrt{5}$.
The corresponding 95\% confidence interval (CI) is estimated as 1.96 $\times$ the standard error.
We perform two-sided paired $t$-tests to evaluate the statistical significance of performance differences between CheXficient and CheXfull, where metrics were computed across repeated experiments.
A $p$-value less than 0.05 indicates a statistically significant difference, whereas a $p$-value greater than or equal to 0.05 suggests comparable performance between them.
For feature-level kNN distance statistics, we use two-sided Welch’s t-test, as the distance distributions involve unequal sample sizes for the subset and full set.

\subsection{Computation resources}

All experiments were conducted on NVIDIA H100 GPUs with 80 GB memory.
CheXficient required approximately 60 H100 GPU-hours for curation and pretraining on 280K paired CXR data, whereas CheXfull required about 220 H100 GPU-hours to train on 1,235K paired CXR data.
For downstream fine-tuning, CheXficient used one H100 GPU for classification and segmentation tasks, and four H100 GPUs for radiology report generation.
All pretraining and fine-tuning procedures were carefully monitored to ensure adequate convergence.

\section{Data availability}
\label{data}

All data used in this study are publicly available: 
MIMIC-CXR (\url{https://physionet.org/content/mimic-cxr-jpg/2.0.0/}); 
CheXpert (\url{https://stanfordmlgroup.github.io/competitions/chexpert/}); 
CheXpert-Plus (\url{https://aimi.stanford.edu/datasets/chexpert-plus});
ReXGradient-160K (\url{https://huggingface.co/datasets/rajpurkarlab/ReXGradient-160K});
PadChest (\url{https://bimcv.cipf.es/bimcv-projects/padchest/});
BIMCV-COVID19 (\url{https://bimcv.cipf.es/bimcv-projects/bimcv-covid19/});
CANDID-PTX (\url{https://figshare.com/articles/dataset/CANDID-PTX/14173982});
CASIA-CXR (\url{https://www.casia-cxr.net/});
Open-I (\url{https://openi.nlm.nih.gov/faq});
NIH ChestX-ray14 (\url{https://nihcc.app.box.com/v/ChestXray-NIHCC/folder/36938765345}); 
BRAX (\url{https://physionet.org/content/brax/1.1.0/});
VinDr-CXR (\url{https://vindr.ai/cxr}); 
VinDr-PCXR (\url{https://vindr.ai/datasets/pediatric-chest-x-ray}); 
ChestDR (\url{https://springernature.figshare.com/articles/dataset/ChestDR_Thoracic_Diseases_Screening_in_Chest_Radiography/22302775}); 
SIIM-PTX (\url{www.kaggle.com/c/siim-acr-pneumothorax-segmentation}); 
Pneumonia2017 (\url{https://data.mendeley.com/datasets/rscbjbr9sj/2}); 
TBX11K (\url{https://datasetninja.com/tbx-11k}); 
JSRT (\url{http://db.jsrt.or.jp/eng.php});
VinDr-RibCXR (\url{https://vindr.ai/ribcxr}).
% CXR-LT (\url{https://physionet.org/content/cxr-lt-iccv-workshop-cvamd/2.0.0/}).

\section{Code availability}
\label{code}

We have publicly released the data curation, pretraining, and evaluation code, along with the pretrained model weights, on GitHub at \url{https://github.com/cwangrun/CheXficient}.
All experimental procedures are described in detail in the Methods section to support independent replication.
To facilitate broader adoption, we also provide standardized downstream modeling tasks that are accessible to a wide scientific audience.
Together, these resources are made available to ensure transparency and to promote further research in this area.

% The code used to train and evaluate CheXficient is available on GitHub \url{https://github.com/cwangrun/CheXficient} 

% All pretraining and evaluation data, model weights, evaluation protocols, and codes are publicly released at

\section*{Acknowledgements}
A.S.C. receives research support from the National Institutes of Health (grants - R01 HL167974, R01 AR077604, R01 EB002524, R01 AR079431, P41 EB027060, and contracts 75N92020C00008, 75N92020C00021); and from GE Healthcare, Philips, Amazon, Microsoft/OpenAI, and Stability.ai. 

% C.B. receives research support from the Promedica Foundation, Chur, Switzerland. Research reported in this publication was made possible in part by the National Institute of Biomedical Imaging and Bioengineering (NIBIB) of the National Institutes of Health which supports the Medical Imaging and Data Resource Center under contracts 75N92020C00008 and 75N92020C00021, and by grant #1R18HS028955 from the Agency for Health Research and Quality.

\section*{Author contributions}

Conceptualization: C.W., Y.Z., A.S.C., C.P.L.;
Methodology: C.W., Y.Z. Y.G., M.V., J.Liu. A.S.C., C.P.L.;
AI model and code development: C.W., Y.Z.;
Dataset development: C.W., Y.Z., Y.G.;
Data analysis: C.W., Y.Z., Y.G., C.M., J.Long.;
Writing - original draft, review, and editing: C.W., Y.Z. Y.G., M.V., C.M., J.Liu., F.P, J.Long., J.B.D, S.G., A.S.C., C.P.L.;
Supervision: S.G., A.S.C., C.P.L.;
Project administration: C.W., Y.Z., A.S.C., C.P.L.;
Funding acquisition: A.S.C., C.P.L.

% C.W. and Y.Z. designed the study and carried out the data collection, data analysis, model construction, and benchmark design. Z.C., M.V., M.P., D.V.V., and J.B.D. carried out the technical model evaluation. A.S.C., S.G., D.V.V., Z.C., J.X., M.V., J.B.D., and C.P.L. designed the clinical reader study. J.X. and Z.C. implemented the reader study. J.X., Z.C., M.V., A.Y., C.O., A.J., S.A., M.S.E.M., E.P.R., E.B.T., C.B., C.F.B, and S.G. carried out the reader study and interpreted the results. Z.C., M.V., J.X., M.P., D.V.V., A.Y., C.B., L.B., J.M.J.V., E.P.R., J.P.C., T.M.A, J.J., J.B.D., A.S.C., and C.P.L. contributed to the technical discussions. All authors contributed to the drafting and revision of the manuscript. 
% J.B.D., A.S.C., and C.P.L. supervised and guided the research.

\section*{Declaration of interests}

Jean-Benoit Delbrouck is an employee of hoppr.ai.

% Some journals require declarations to be submitted in a standardised format. Please check the Instructions for Authors of the journal to which you are submitting to see if you need to complete this section. If yes, your manuscript must contain the following sections under the heading `Declarations':

% \begin{itemize}
% \item Conflict of interest/Competing interests (check journal-specific guidelines for which heading to use)
% \item Ethics approval and consent to participate
% \item Consent for publication
% \item Materials availability
% \end{itemize}

\newpage

\renewcommand{\tablename}{Extended Data Table}
\setcounter{table}{0}
\begin{table*}[h]
\centering
\setlength{\extrarowheight}{0pt}
\addtolength{\extrarowheight}{\aboverulesep}
\addtolength{\extrarowheight}{\belowrulesep}
\setlength{\aboverulesep}{0pt}
\setlength{\belowrulesep}{0pt}
\caption{Comparative overview of CheXficient with other large-scale pretrained medical foundation models.
}
\vspace{-6pt}
\label{tab:comparative overview}

\setlength{\tabcolsep}{1.3 mm}
\resizebox{1.0\linewidth}{!}{

\begin{tabular}{lcccccc} 
\hline
Model                                          & Image encoder & Parameter size & Image resolution & Pretraining data size & Pretraining type & Data accessibility  \\ 
\hline
CheXficient                                    & ViT-Base      & 86 M           & $378 \times 378$ & 280 K                 & image-text       & public              \\
CheXZero~\cite{tiu2022expert}                  & ViT-Base      & 86 M           & $224 \times 224$ & 377 K                 & image-text       & public              \\
BioViL-T~\cite{bannur2023learning}             & ResNet-50     & 25.6 M         & $512 \times 512$ & 377 K                 & image-text       & public              \\
LLaVA-Rad~\cite{zambrano2025clinically}        & ViT-Base      & 86 M           & $518 \times 518$ & 697 K                 & image-text       & public              \\
RAD-DINO~\cite{perez2025exploring}             & ViT-Base      & 86 M           & $518 \times 518$ & 883 K                 & self-supervised  & public + private    \\
CheXagent~\cite{chen2024chexagent}             & ViT-Large     & 307 M          & $512 \times 512$ & 1,077 K               & image-text       & public              \\
Libra~\cite{zhang2025libra}                    & ViT-Base      & 86 M           & $518 \times 518$ & 1,213 K               & self-supervised  & public              \\
MAIRA-2~\cite{bannur2024maira}                 & ViT-Base      & 86 M           & $518 \times 518$ & 1,404 K               & self-supervised  & public + private    \\
BiomedCLIP~\cite{zhang2023biomedclip}          & ViT-Base      & 86 M           & $224 \times 224$ & 15 M                  & image-text       & private             \\
RadFM~\cite{wu2025towards}                     & 12-layer 3D ViT  & 90 M        & $512 \times 512$ & 16 M                  & image-text       & public + private    \\
MedVersa~\cite{zhou2024medversa}               & Swin Transformer  & 88 M       & $224 \times 224$ & 29 M                  & image-text       & public              \\
MedGemma~\cite{sellergren2025medgemma}         & SoViT-400m    & 400 M          & $448 \times 448$ & 33 M                  & image-text       & public + private    \\
\hline
\end{tabular}

}
\end{table*}
\clearpage

\renewcommand{\figurename}{Extended Data Fig.}
\setcounter{figure}{0}
\begin{figure}[!t] 
\centering
\includegraphics[width=1.0\linewidth]{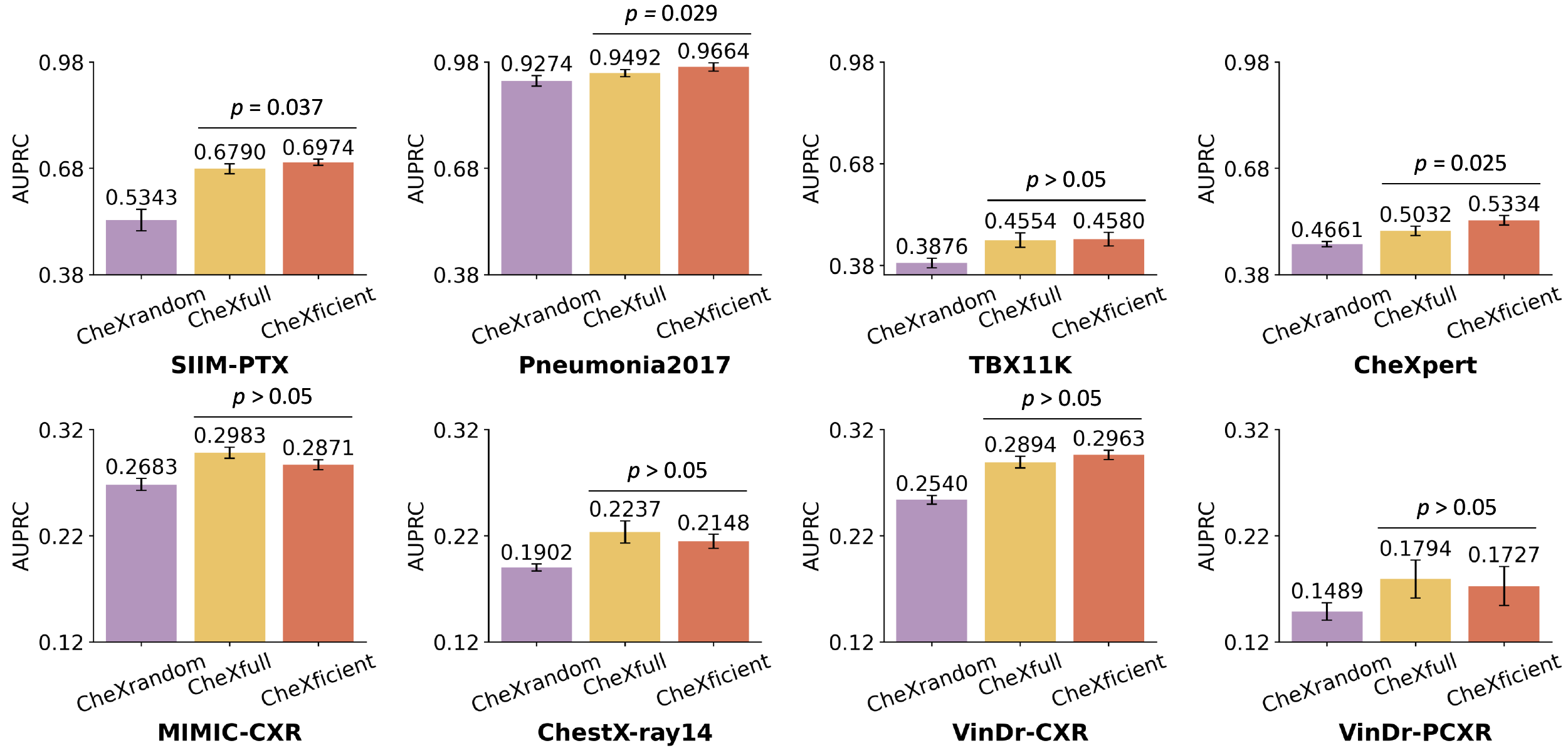}
\vspace{-15 pt}
\caption{Performance (AUPRC) on zero-shot findings classification. 
We evaluate the pretrained models on 8 public datasets. 
Among them, SIIM-PTX, Pneumonia2017, and TBX11K are from external domains unseen in pretraining, while the remaining 5 datasets are used for internal evaluation.
Compared to CheXfull, CheXficient achieves higher AUROC on 3 datasets ($p < 0.05$), and comparable performance on 5 datasets ($p >$ 0.05).
For each dataset, CheXficient outperforms CheXrandom by large margins. 
We present the mean $\pm$ 95\% confidence interval (CI) of AUPRC for each model. The listed $p$-values are calculated using two-sided $t$-tests.
}
\label{fig:zero-shot classification AUPRC}
% \vspace{-10pt}
\end{figure}
\clearpage

\begin{figure}[!t] 
\centering
\includegraphics[width=1.0\linewidth]{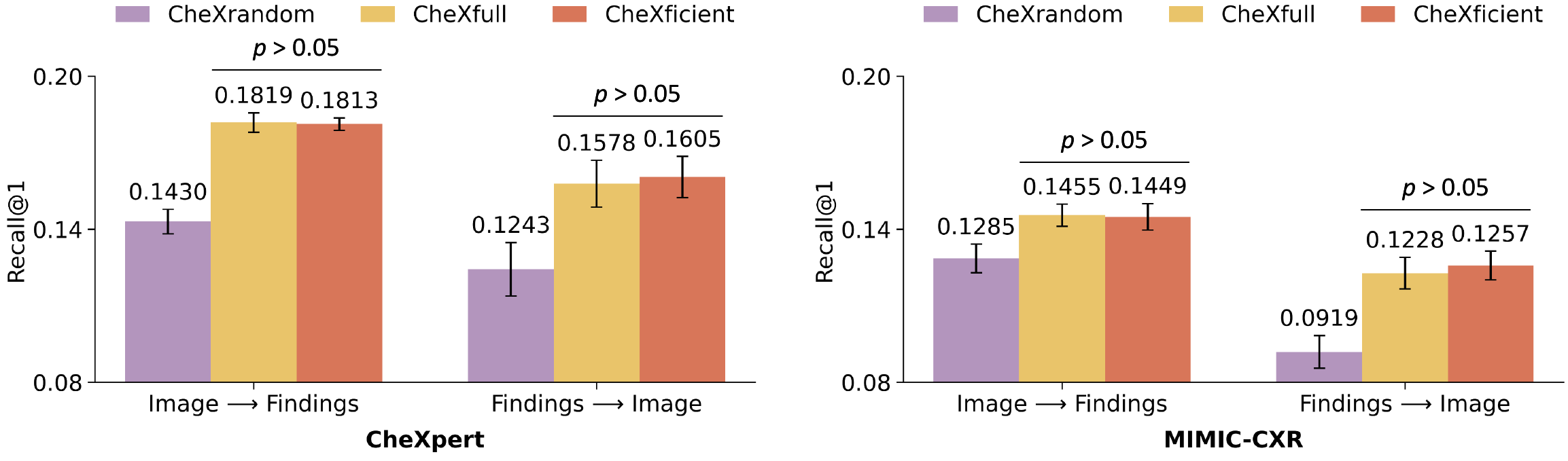}
\vspace{-15 pt}
\caption{Performance on zero-shot cross-modal retrieval (Image $\rightarrow$ Findings and Findings $\rightarrow$ Image).
We evaluate the pretrained models on the public CheXpert and MIMIC-CXR benchmarks. 
CheXficient achieves performance comparable to CheXfull on both datasets ($p > 0.05$), while outperforming CheXrandom.
We report the mean $\pm$ 95\% CI of Recall@1 (The recall of retrieving the exact paired CXR Findings section (or image) within the top-1 result for a given CXR image (or Findings section)). 
% The listed $p$ values are calculated using two-sided t-test.
% \chong{show Findings and Impressions instead?}
}
\label{fig:zero-shot retrieval Findings}
% \vspace{-10pt}
\end{figure}
\clearpage

\begin{figure}[!t] 
\centering
\includegraphics[width=1.0\linewidth]{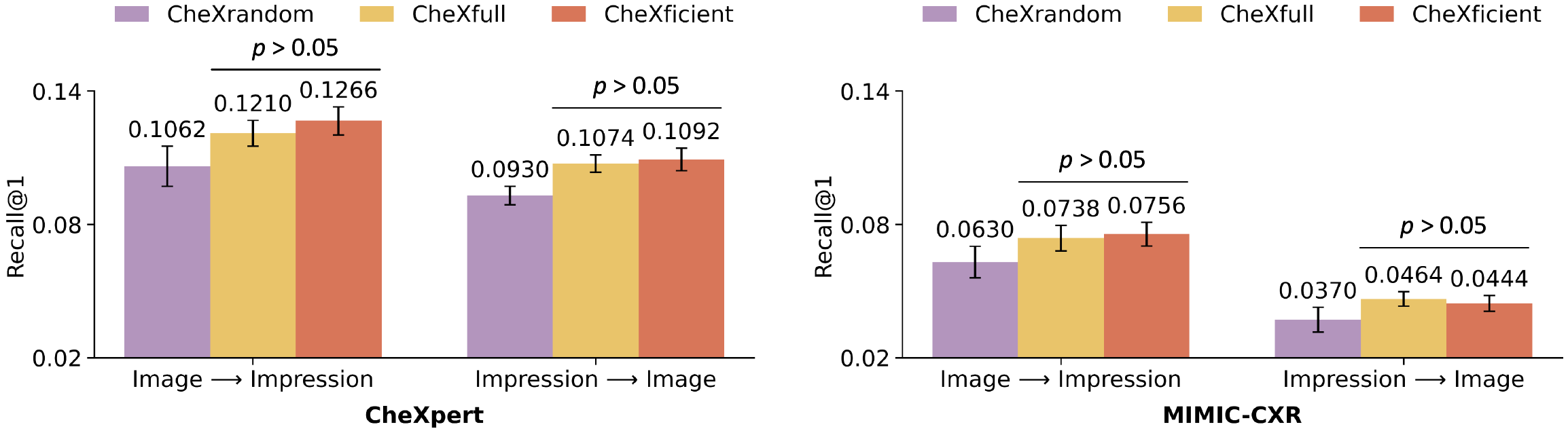}
\vspace{-15 pt}
\caption{Performance on zero-shot cross-modal retrieval (Image $\rightarrow$ Impression and Impression $\rightarrow$ Image).
We evaluate the pretrained models on the public CheXpert and MIMIC-CXR benchmarks. 
CheXficient achieves performance comparable to CheXfull on both datasets ($p > 0.05$), while outperforming CheXrandom.
We report the mean $\pm$ 95\% CI of Recall@1 (The recall of retrieving the exact paired CXR Impression section (or image) within the top-1 result for a given CXR image (or Impression section)). 
% The listed $p$-values are calculated using two-sided $t$-tests.
% \chong{show Findings and Impressions instead?}
}
\label{fig:zero-shot retrieval Impression}
% \vspace{-10pt}
\end{figure}
\clearpage

\begin{figure}[!t] 
\centering
\includegraphics[width=1.0\linewidth]{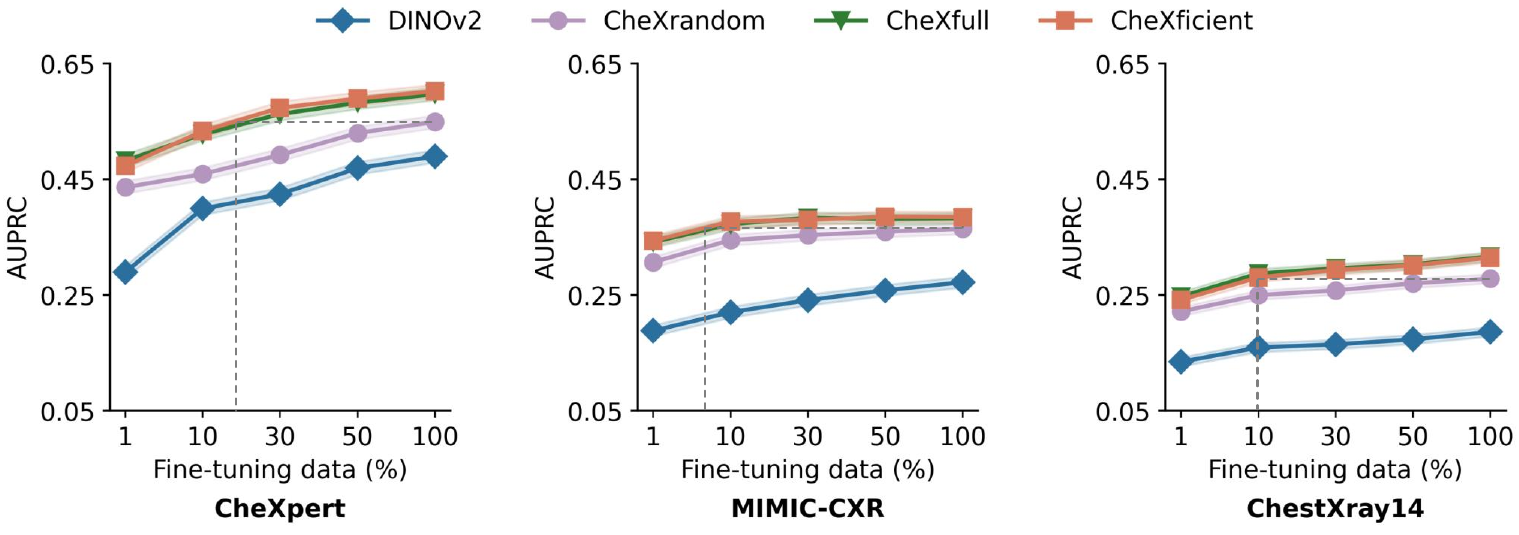}
\vspace{-15 pt}
\caption{
Performance (AUPRC) comparison of CheXficient with other pretrained models in label-efficient disease prediction across three representative datasets: CheXpert ($n$ = 500, 14 categories), MIMIC-CXR ($n$ = 3,082, 14 categories), and ChestXray14 ($n$ = 25,596, 15 categories). 
Performance is reported at varying proportions of labeled training data. 
Vertical dashed lines indicate the amount of labeled data required for CheXficient to match the performance of its counterpart CheXrandom.
}
\label{fig:classification label efficiency AUPRC}
% \vspace{-10pt}
\end{figure}
\clearpage

\begin{algorithm}[!tb]
% \setlength{\algomargin}{0.1em}
% \SetAlgoInsideSkip{0.9em}
\caption{Data-curated Pretraining Procedure of CheXficient}
\label{alg:training_algorithm}
\KwData{Pretraining set $\mathcal{D} = \{ ({x}_i^{\text{img}}, {x}_i^{\text{txt}}) \}_{i=1}^{|\mathcal{D}|}$, number of training iterations $T$ for curation, super-batch size $|\mathcal{S}|$, mini-batch size $|\mathcal{M}|$.}
\KwResult{Image encoder $\theta_{\text{img}}$, text encoder $\theta_{\text{txt}}$, curation set $\mathcal{D}_c$.
% and prototypes $\{\mathbf{p}_k\}_{k=1}^{K}$ 
}

Initialize $\mathcal{D}_c = \emptyset$, initialize prototypes $\{\mathbf{p}_k\}_{k=1}^{K}$ by k-means;

\For{iteration $t = 1$ \KwTo $T$}{

  Given a large super-batch $\mathcal{S}$ sampled from $\mathcal{D}$; 
  
  Extract unified embeddings $\mathbf{z}_i = {\mathsf{concat}}(\mathbf{z}_i^{\text{img}}, \mathbf{z}_i^{\text{txt}})$ for each sample in $\mathcal{S}$; 
  
  Compute the distance of each $\mathbf{z}_i$ to its nearest prototype: $\min_k \, \mathsf{dist}(\mathbf{z}_i, \mathbf{p}_k)$; 
  
  Select data based on sorted distances and form the mini-batch $\mathcal{M}$; 
  
  Contrastive pretraining on $\mathcal{M}$ with InfoNCE loss; 
  
  Calculate prototypes based on embeddings of the selected samples in $\mathcal{M}$; 

  Update prototypes $\{\mathbf{p}_k \}_{k=1}^{K}$ via exponential moving average; 

  Add selected samples into the curation set: $\mathcal{D}_c = \mathcal{D}_c \cup \mathcal{M}$.
}

\end{algorithm}
\clearpage

\begin{appendices}

\section{Additional analysis}
\label{Additional_analysis}

To further investigate the effects of data curation in pretraining, we additionally analyze patient demographic information (e.g., age and race), distribution of frontal and lateral CXR scans, and distribution of token length of CXR reports. 

\subsection{Patient age}
\label{age}

\begin{figure}[!t] 
\centering
\includegraphics[width=0.72\linewidth]{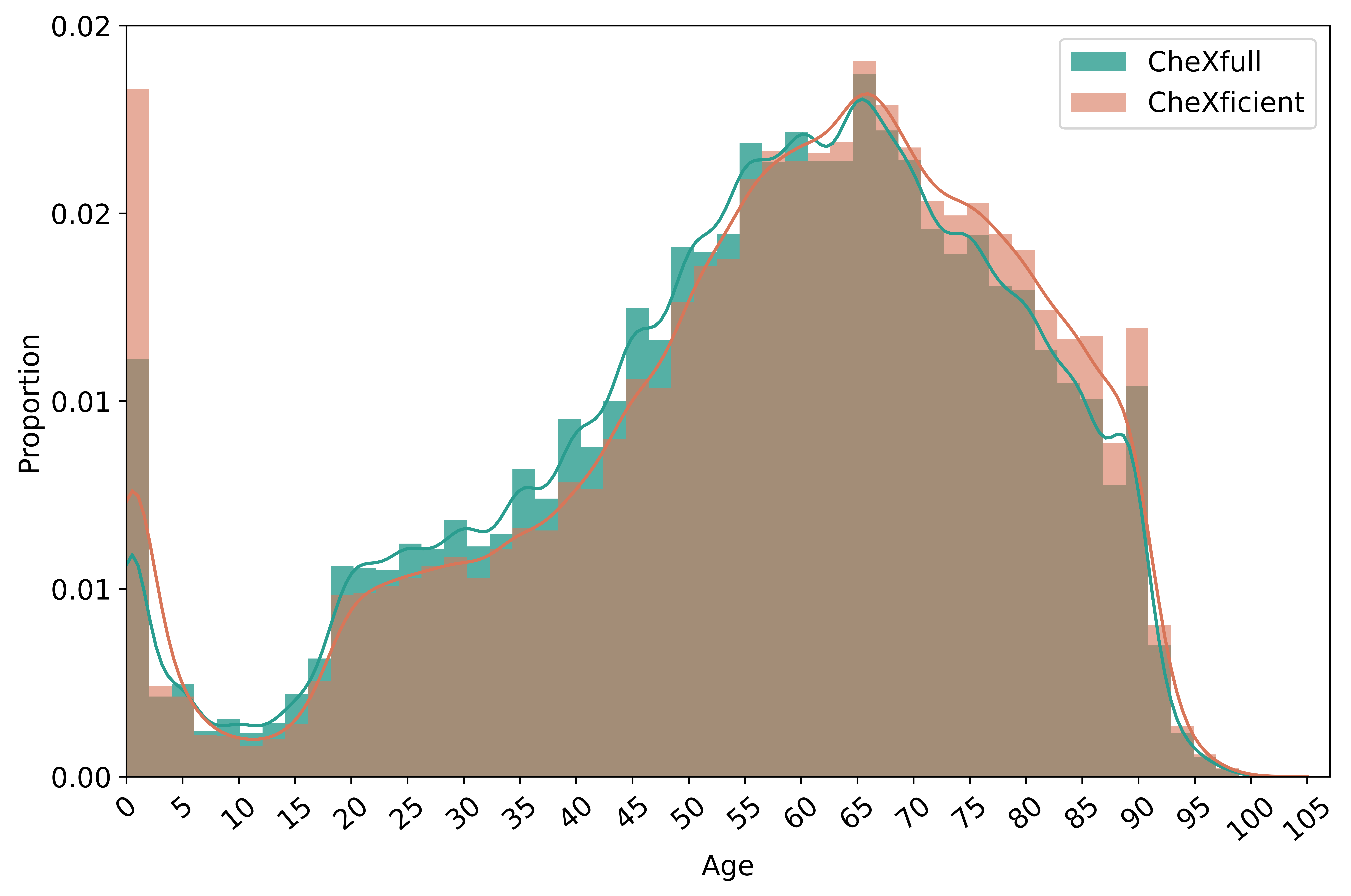}
\vspace{-1 pt}
\caption{
Distribution of patient age in the full training set (CheXfull) and the curated subset (CheXficient).
The curated subset includes relatively more samples from two age groups: pediatric patients (0–5 years old) and older adults (over 65 years old).
}
\label{fig:age}
% \vspace{-10pt}
\end{figure}

Patient demographic information is an important factor to consider when developing medical AI models. 
We examine how data curation during pretraining affects the distribution of patient age by comparing the full training set and the curated subset.
Figure~\ref{fig:age} shows that the curated subset tends to include relatively more samples from two age groups: pediatric patients (0–5 years old) and older adults (over 65 years old).
This observation may be relevant because pediatric CXRs exhibit greater anatomical variability, and CXRs of older adults often show a wider range of age-related pathologies due to their higher prevalence in this population.
Hence, the curated pretraining data emphasizes samples from these two age groups.

\renewcommand{\figurename}{Figure}

\subsection{Patient race}
\label{race}

\begin{figure}[!t] 
\centering
\includegraphics[width=0.76\linewidth]{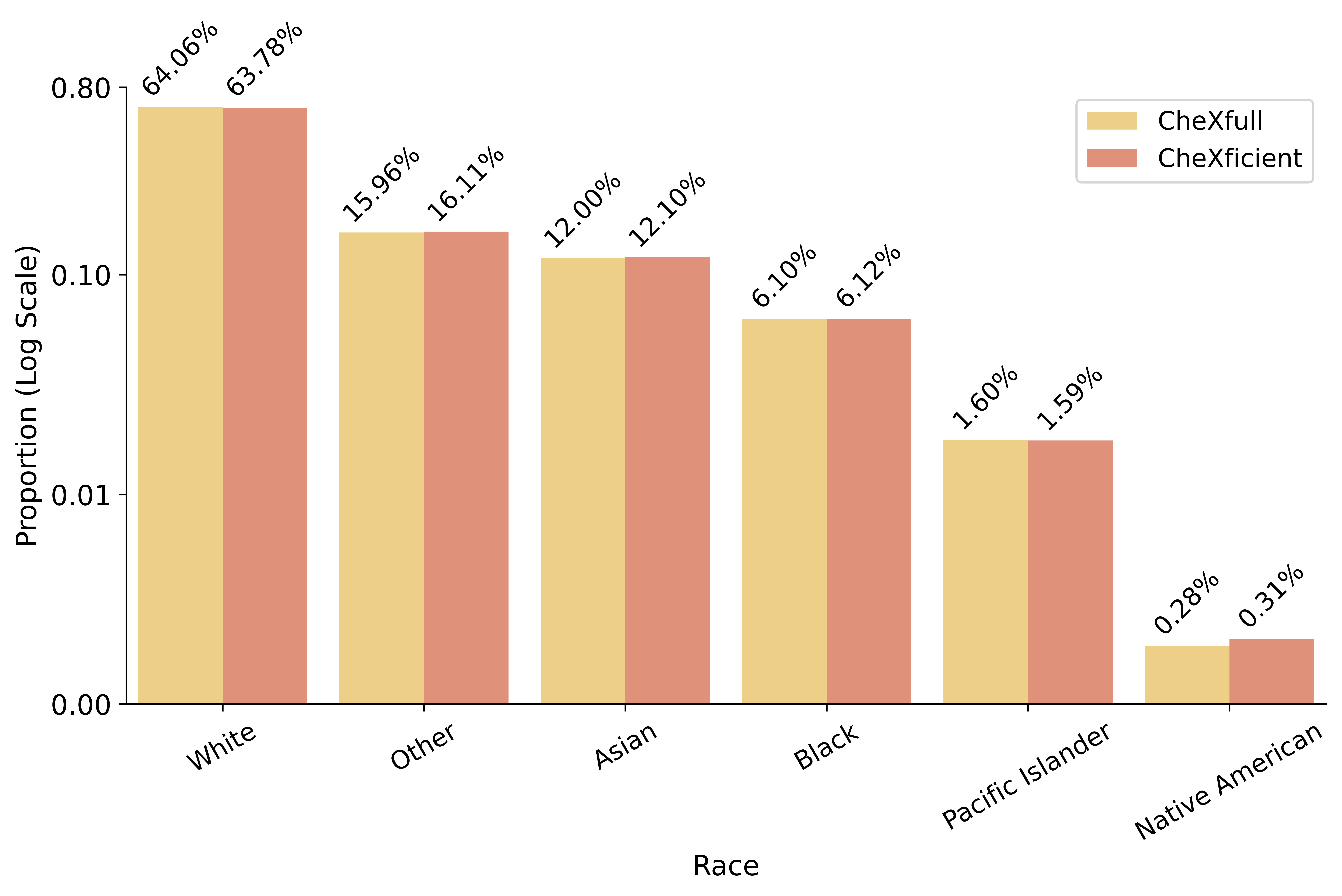}
\vspace{-1 pt}
\caption{
Distribution of patient race in the full training set (CheXfull) and the curated subset (CheXficient). 
The two sets show similar race distributions.
}

\label{fig:race}
% \vspace{-10pt}
\end{figure}

Understanding the distribution of patient race in the training data helps assess whether curated pretraining could introduce unintended biases.
We examine the distribution of patient race in the CheXpert dataset, which includes `White', `Asian', `Black', `Pacific Islander', `Native American', and `Other'.
Figure~\ref{fig:race} shows that the distribution of races is similar between the full training set and the curated subset.
This observation indicates that the curation process does not appear to be associated with patient race.
This is consistent with prior work~\cite{duffy2022confounders,perez2025exploring}, which points out that there is little detectable information about race and ethnicity in chest X-ray features.

\renewcommand{\figurename}{Figure}

\subsection{Frontal and lateral CXR scans}
\label{frontal_lateral}

\begin{figure}[!t] 
\centering
\includegraphics[width=0.53\linewidth]{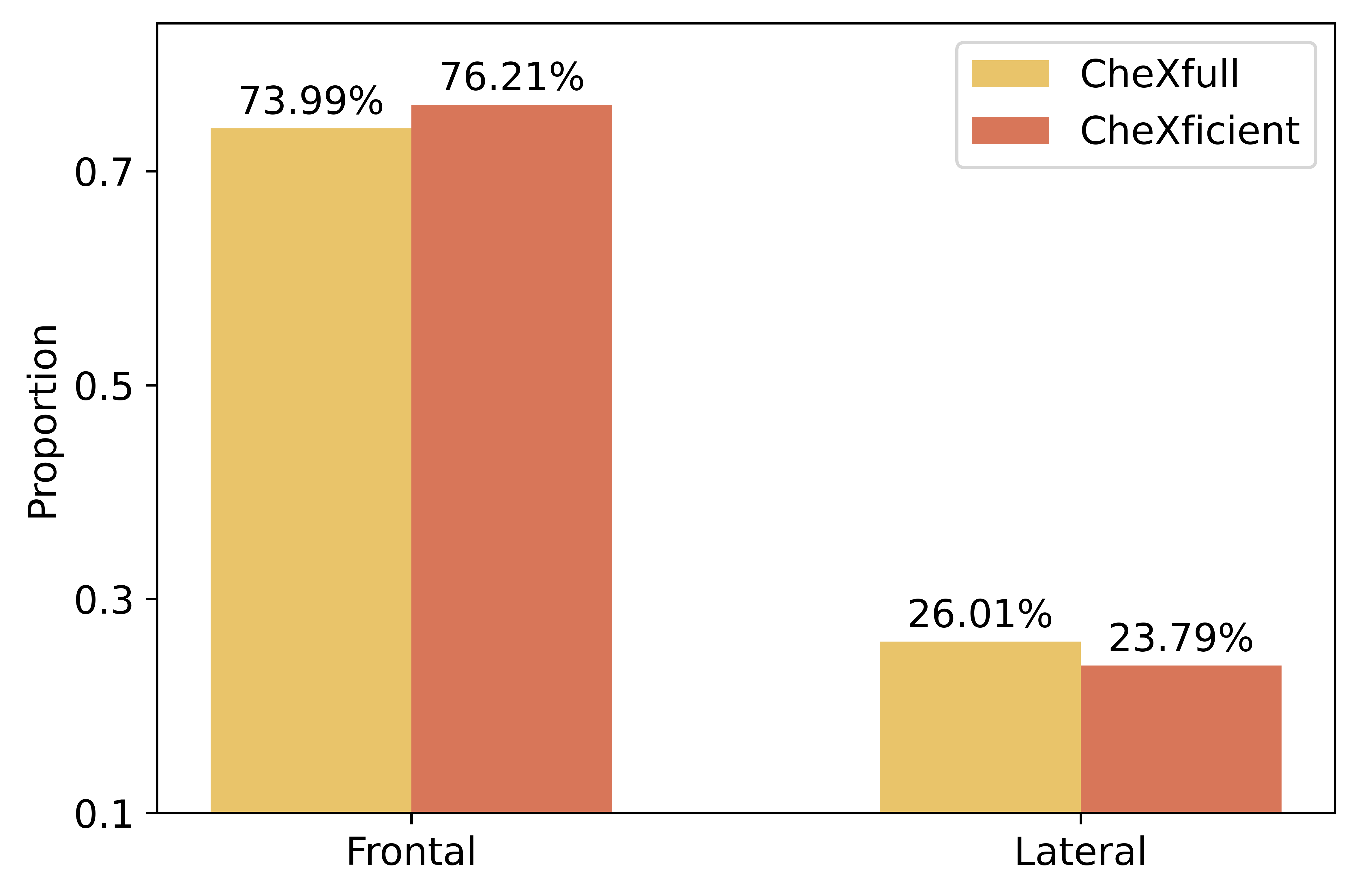}
\vspace{-1 pt}
\caption{
Distribution of frontal and lateral scans in the full training set (CheXfull) and the curated subset (CheXficient).
The curated subset contains a higher proportion of frontal scans in comparison with the full training set.
}
\label{fig:frontal_lateral}
% \vspace{-10pt}
\end{figure}

Frontal and lateral chest X-rays (CXRs) are routinely acquired in clinical practice and are often jointly used to disambiguate radiological findings.
We examine whether CheXficient exhibits a preference for either view during data curation.
As shown in Figure~\ref{fig:frontal_lateral}, the curated subset contains a slightly higher proportion of frontal scans than the original dataset.
One possible explanation is that frontal CXRs may provide more informative visual cues for many radiological findings, whereas certain findings are less clearly visible on lateral scans~\cite{hashir2020quantifying}.
This observation suggests that the curation process tends to prioritize the informative frontal scans during pretraining.

\renewcommand{\figurename}{Figure}

\subsection{Token length of CXR reports}
\label{text_token}

\begin{figure}[!t] 
\centering
\includegraphics[width=0.72\linewidth]{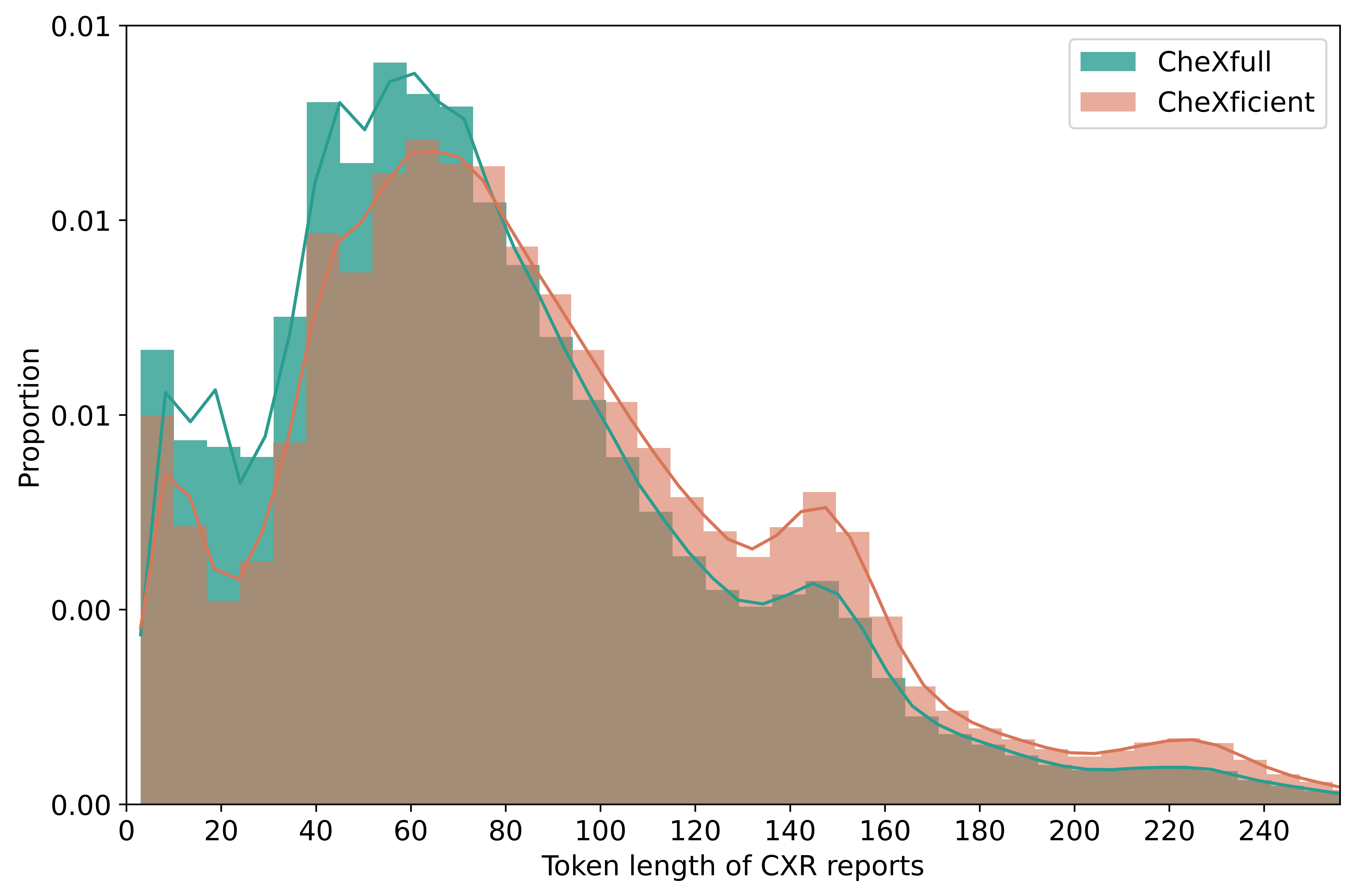}
\vspace{-1 pt}
\caption{
Distribution of CXR report token lengths in the full training set (CheXfull) and the curated subset (CheXficient).
The curated subset demonstrates a higher proportion of longer CXR reports with high informativeness.
}
\label{fig:text_token}
% \vspace{-10pt}
\end{figure}

We further analyze the token length of CXR reports as a proxy for the informativeness of training samples. 
In general, CXR reports describing normal findings tend to be shorter, whereas those containing rich pathological findings require more detailed descriptions.
As shown in Figure~\ref{fig:text_token}, the curated subset exhibits a relatively higher proportion of longer CXR reports, 
suggesting that the data curation process systematically favors samples with more informative clinical descriptions.

\renewcommand{\figurename}{Figure}

\section{Ablation study}
\label{ablation}

To showcase the advantages of the CheXficient framework, we conduct ablation studies to highlight its multimodal embedding design (Section~\ref{ablation_embeddings}).
We further evaluate its robustness through a sensitivity analysis with respect to the number of prototypes (Section~\ref{ablation_prototype_number}).

\subsection{Multimodal embedding vs. unimodal embedding}
\label{ablation_embeddings}

In CheXficient, data curation can be performed using either unified multimodal embeddings obtained by concatenating image and text features, i.e., ${\mathsf{concat}}(\mathbf{z}_i^{\mathrm{img}}, \mathbf{z}_i^{\mathrm{txt}})$, or unimodal embeddings derived from images ($\mathbf{z}_i^{\mathrm{img}}$) or reports ($\mathbf{z}_i^{\mathrm{txt}}$) alone.
Table~\ref{tab:multimodal embedding} presents an ablation study comparing these choices.
While no consistent trend is observed indicating the superiority of a single modality, combining image and text embeddings generally yields better performance than using either modality alone.
This suggests that paired image-report data provide richer, complementary semantic cues that more accurately reflect data importance for effective data selection, ultimately contributing to improved representation learning.

\renewcommand{\tablename}{Table}
\begin{table*}[h]
\centering
\setlength{\extrarowheight}{0pt}
\addtolength{\extrarowheight}{\aboverulesep}
\addtolength{\extrarowheight}{\belowrulesep}
\setlength{\aboverulesep}{0pt}
\setlength{\belowrulesep}{0pt}
\caption{
Effect of different feature embeddings used in CheXficient pretraining, e.g., image-only $\mathbf{z}_i^{\mathrm{img}}$, text-only $\mathbf{z}_i^{\mathrm{txt}}$, and image-text concatenation ${\mathsf{concat}}(\mathbf{z}_i^{\mathrm{img}}, \mathbf{z}_i^{\mathrm{txt}})$, on 280K curated CXR data pairs.
Zero-shot findings classification performance (AUROC) is reported across eight datasets.
}
\vspace{-6pt}
\label{tab:multimodal embedding}

\setlength{\tabcolsep}{2.0 mm}
\resizebox{1.0\linewidth}{!}{

\begin{tabular}{lccc} 
\toprule
Dataset       & $\mathbf{z}_i^{\text{img}}$                       & $\mathbf{z}_i^{\text{txt}}$                   & ${\mathsf{concat}}(\mathbf{z}_i^{\text{img}}, \mathbf{z}_i^{\text{txt}})$               \\ 
\midrule
SIIM-PTX      & 0.8920 [0.8858, 0.8981] & 0.8937 [0.8900, 0.8975]          & \textbf{0.9074 [0.9001, 0.9148]}  \\
Pneumonia2017 & 0.9481 [0.9416, 0.9546] & 0.9490 [0.9447, 0.9532]          & \textbf{0.9659 [0.9597, 0.9722]}  \\
TBX11K        & 0.8209 [0.8145, 0.8274] & 0.8174 [0.8110, 0.8238]          & \textbf{0.8248 [0.8183, 0.8313]}  \\
CheXpert      & 0.8407 [0.8361, 0.8453] & 0.8335 [0.8290, 0.8380]          & \textbf{0.8535 [0.8501, 0.8569]}  \\
MIMIC-CXR     & 0.6613 [0.6558, 0.6669] & 0.6758 [0.6708, 0.6807]          & \textbf{0.6866 [0.6824, 0.6908]}  \\
ChestX-ray14  & 0.7191 [0.7166, 0.7216] & \textbf{0.7312 [0.7255, 0.7369]} & 0.7308 [0.7281, 0.7336]           \\
VinDr-CXR     & 0.8966 [0.8929, 0.9003] & 0.8927 [0.8866, 0.8989]          & \textbf{0.9005 [0.8929, 0.9081]}  \\
VinDr-PCXR    & 0.6992 [0.6915, 0.7070] & 0.6956 [0.6882, 0.7031]          & \textbf{0.7019 [0.6951, 0.7087]}  \\
\bottomrule
\end{tabular}

}
\end{table*}

\subsection{Effect of the number of prototypes}
\label{ablation_prototype_number}

\begin{figure}[!t] 
\centering
\includegraphics[width=0.75\linewidth]{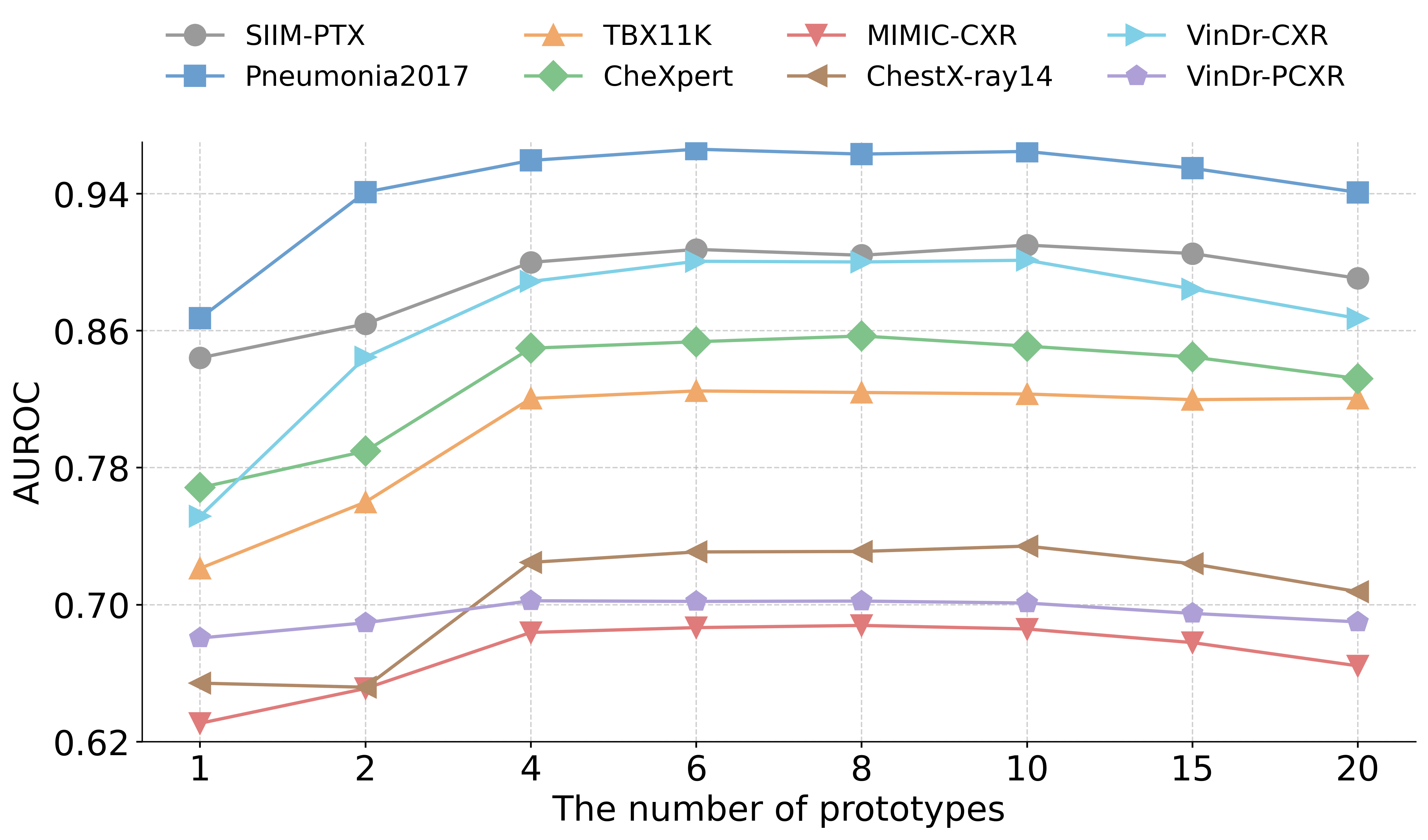}
\vspace{-1 pt}
\caption{
Effect of the number of prototypes used in CheXficient pretraining on zero-shot findings classification performance across eight datasets.
}
\label{fig:prototypenumber}
% \vspace{-10pt}
\end{figure}

We further analyze the sensitivity of CheXficient to the number of prototypes $K$, as shown in Figure~\ref{fig:prototypenumber}, by varying $K$ under the same pretraining data budget (i.e., 280K data pairs).
Overall, the zero-shot findings classification performance is relatively robust when $K$ ranges from 4 to 10.
Using too few prototypes fails to adequately capture the complexity and diversity of the training data, leading to sub-optimal performance.
In contrast, employing an excessive number of prototypes introduces surplus or even harmful information, resulting in performance degradation and increased model complexity.
Based on these observations, we set the number of prototypes to $K=6$ for all remaining experiments.

\section{Downstream implementation details}
\label{downstream details}

The implementation details for training and evaluating each downstream model are described below.

\subsection{Image classification}

All downstream image classification experiments are conducted on a single NVIDIA H100 GPU.
We use a batch size of 512 and optimize the models with AdamW, using a base learning rate of $3 \times 10^{-2}$, a weight decay of $1 \times 10^{-4}$, and a step-based learning rate decay scheduler.
For image pre-processing, we apply center cropping 
% and resizing to $378 \times 378$.
and no additional image augmentation is applied during downstream training.
CXR image intensities are normalized using a mean of [0.4814, 0.4578, 0.4082] and a standard deviation of [0.2686, 0.2613, 0.2757].
All classification models are trained for 15 epochs.
The final checkpoint is used for inference on the test set, as no overfitting is observed based on the validation loss.
We conduct experiments over five runs with different random seeds.
Under this GPU setup, fine-tuning a linear classification head on top of the visual encoder takes approximately 0.41 seconds per iteration.

\subsection{Semantic segmentation}

We evaluate the segmentation tasks on 1 NVIDIA H100 GPU.
We use a training batch size of 256, AdamW optimizer, a base learning rate of $3 \times 10^{-4}$, a weight decay of $1 \times 10^{-4}$, and a step-based learning rate decay scheduler.
For image pre-processing, we apply center cropping and 
% and resizing to $378 \times 378$.
no additional image augmentation is applied during downstream training.
CXR image intensities are normalized using a mean of [0.4814, 0.4578, 0.4082] and a standard deviation of [0.2686, 0.2613, 0.2757].
The segmentation models are trained for a total of 120 epochs, using a combined Dice loss and cross-entropy loss as the training objective.
For lung segmentation on the JSRT dataset, we use a data split of 172/25/50 for the training, validation, and test sets, respectively, and report Dice scores on the test set.
For pneumothorax segmentation on SIIM-PTX and rib segmentation on VinDr-RibCXR, we follow the official data splits provided by the datasets.
The model with the lowest validation loss is selected for evaluation on the test set.
Under the given GPU setup, training a U-Net decoder as the segmentation head on top of the visual encoder takes approximately 0.45 seconds per iteration.

\subsection{Radiology report generation}

All radiology report generation experiments are conducted on 1 compute node equipped with 4 NVIDIA H100 GPUs (80GB each).
We follow the same training hyper-parameters as LLaVA-Rad~\cite{zambrano2025clinically}.
Specifically, we use a total batch size of 128 (32 per GPU), a cosine learning rate scheduler with a warmup phase over 3\% of the training steps, and a base learning rate of $3 \times 10^{-5}$.
We adopt a two-stage fine-tuning strategy over three epochs, where the image encoder is kept frozen throughout both stages.
In the first stage (feature alignment), both the image encoder and the large language model (LLM) are frozen, and only the MLP projection layer is trained.
In the second stage (end-to-end fine-tuning), the MLP projection layer and the LLM are jointly updated, while the image encoder remains frozen.
Training on MIMIC-CXR and ReXGradient-160K requires approximately 0.5 and 0.2 days, respectively.

% Finally, we use 32-bit full precision for decoding up to 150 tokens with a batch size of 1 during inference.

\end{appendices}


\clearpage

\bibliography{my-bib}% common bib file

\end{document}